%% file: main.tex
\documentclass[10pt,twocolumn,letterpaper]{article}

\usepackage[camera]{cvpr}      
\usepackage{times}
\usepackage{epsfig}
\usepackage{graphicx}
\usepackage{amsmath}
\usepackage{amssymb}
\usepackage{booktabs}
\usepackage{multirow}
\usepackage{enumitem}
\usepackage{xcolor}
\usepackage{caption}
\usepackage{subcaption}
\usepackage{algorithm}
\usepackage{algpseudocode}
\usepackage{mathtools}
\usepackage{float}
\usepackage{yhmath}
\usepackage[accsupp]{axessibility}  

\input{preamble}

\definecolor{my_blue}{RGB}{44,115,182}

\definecolor{my_red}{RGB}{182,60,44}

\definecolor{my_green}{RGB}{76,187,23}

\definecolor{cvprblue}{rgb}{0.21,0.49,0.74}
\usepackage[pagebackref,breaklinks,colorlinks,allcolors=cvprblue]{hyperref}

\definecolor{total_red}{RGB}{225,10,10}

\def\paperID{40212} 
\def\confName{CVPR}
\def\confYear{2026}

\title{VectorArk: Learning Practical Image Vectorization with Rounded Polygon Representation}

\author{Tarun Gehlaut \quad Difan Liu \quad Charu Bansal \quad Krutik Malani\\
Souymodip Chakraborty \quad Ankit Phogat \quad Matthew Fisher \quad Vineet Batra\\[0.5em]
Adobe\\
}

\begin{document}

\twocolumn[{
\renewcommand\twocolumn[1][]{#1}
\maketitle
    \vspace{-1.7em}
    \centering
    \includegraphics[width=0.89\textwidth]{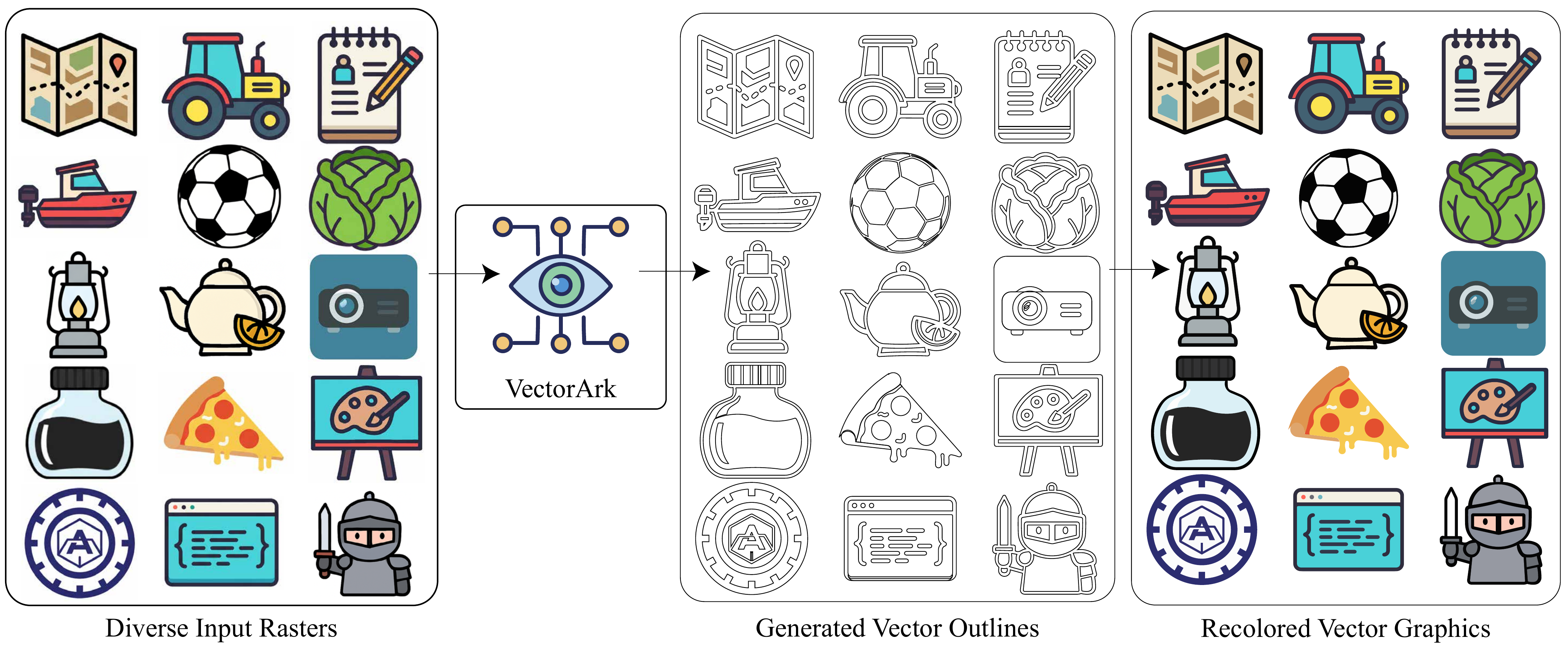}
    \captionsetup{type=figure}
    \captionof{figure}{Our framework handles real-world vectorization scenarios, including degraded outputs from text-to-image models and low-resolution vector renditions. By decoupling geometry prediction from styling through a specialized rounded-polygon representation and outline-based pipeline, our method generates clean, smooth vector primitives that are subsequently stylized in a post-processing step. }
    \vspace{0.7em}
    \label{fig:teaser}
}]
\input{sec/0_abstract}    
\input{sec/1_intro}
\input{sec/3_finalcopy}
{
    \small
    \bibliographystyle{ieeenat_fullname}
    \bibliography{main}
}

\input{sec/X_suppl}

\end{document}

%% file: preamble.tex
\newif\ifshowrebuttalmarkup
\showrebuttalmarkuptrue
\showrebuttalmarkupfalse  

\usepackage[normalem]{ulem}

\ifshowrebuttalmarkup
\newcommand{\rebuttal}[1]{{\color{blue}#1}}
\newcommand{\rebuttaldel}[1]{{\color{red}\sout{#1}}}
\else
\newcommand{\rebuttal}[1]{#1}
\newcommand{\rebuttaldel}[1]{}
\fi



%% file: sec/0_abstract.tex
\begin{abstract}
Recent vision-language model (VLM)–based approaches have achieved impressive results on image vectorization tasks. However, they are typically evaluated on synthetic benchmarks, where clean SVGs are rasterized at high resolution and then re-vectorized. As a result, these methods generalize poorly to real-world scenarios, such as images with unknown rasterization methods or those generated by text-to-image models.
We introduce VectorArk, a new VLM-based model designed for robust and practical image vectorization. VectorArk employs a novel rounded polygon representation that simplifies the learning process while naturally producing smooth, visually appealing primitives. We also propose a degradation model that enhances robustness across diverse and imperfect inputs.
Our experiments show that, in contrast to previous methods, VectorArk achieves superior geometric completeness and artifact suppression across multiple datasets, with comprehensive ablations validating the contribution of each component.
\end{abstract}

%% file: sec/1_intro.tex
\section{Introduction}

%
%
%
%
%
%
%
%
%

\begin{figure*}[t]
    \centering
    \includegraphics[width=0.95\textwidth]{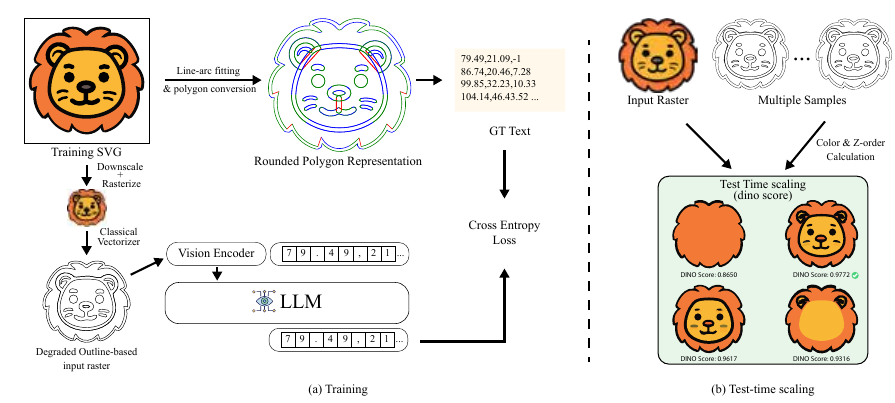}
    \caption{\textbf{Training and inference pipelines.} \emph{(a) Training:} Both paths are used during training. The upper path converts clean SVGs to rounded polygon ground truth tokens via line-arc fitting. The lower path generates degraded outline rasters through downscaling, rasterization, and classical vectorization. The VLM learns to predict clean geometry from degraded inputs using cross-entropy loss. \emph{(b) Test-time scaling:} At inference, input rasters are stochastically decoded into multiple SVG candidates, post-processed for color and z-order, then ranked by DINO similarity to select the best output (checkmark indicates highest score).}
\label{fig:pipeline}
\end{figure*}

Converting raster images to vector graphics is a classical but challenging task in computer graphics. Compared with raster images, vector-based representations offer resolution-independent rendering, compact storage, and improved editability, making them widely used in applications such as vector graphics editors, web rendering engines, and data visualization tools. 
Classic image-to-vector approaches \cite{Potrace,ch25,xu2022live} excel in reliability and computational efficiency, yet they frequently produce imprecise geometry or redundant paths, often lacking semantic awareness of the underlying layer structure.

\begin{figure}[t]
\centering
\includegraphics[width=0.8\columnwidth]{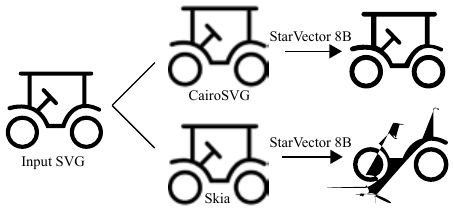}
\caption{\textbf{Sensitivity to rasterization methods.} Given the same input SVG (left), existing methods like StarVector~\cite{Rodriguez_StarVector_2025} produce drastically different outputs when the rasterization backend is changed (CairoSVG ~\cite{cairosvg} vs. Skia ~\cite{skia}). This brittleness demonstrates that prior work is not robust to practical vectorization scenarios where input characteristics vary.}
\label{fig:rasterizer_sensitivity}
\end{figure}

Recent data-driven approaches \cite{Rodriguez_StarVector_2025, yang2025omnisvg, xing2024llm4svg} finetune multimodal LLMs to generate \emph{Scalable Vector Graphics} (SVG) by learning from artist-designed vector data. These methods produce impressive results and are capable of capturing human preferences in control-point placement and layer structuring. 
However, these approaches face several challenges that limit their robustness and generalizability.
First, previous approaches use SVG commands as the underlying representation. While this representation is well suited for visualization, it is neither compact nor canonical from a generative modeling perspective, which complicates the model learning process. 
Second, prior work typically evaluates on clean, noise-free rasters rendered directly from SVG ground truth, but practical image-to-vector tasks demand robustness to diverse input characteristics. As shown in Figure~\ref{fig:rasterizer_sensitivity}, existing methods exhibit unpredictable behavior when the rasterization pipeline is modified—for instance, switching from one rendering backend to another causes significant degradation in output quality. Similarly, outputs from text-to-image models~\cite{gemini2025gemini2_5, openai2024gpt4o} are increasingly used as inputs for vectorization, yet these generated images often contain distorted shapes and visual characteristics that differ substantially from SVG renderings. As a result, previous methods trained on artist-designed vector data struggle to generalize effectively to such real-world scenarios where input characteristics vary.
Third, the stochastic nature of autoregressive models can cause them to succeed or fail unpredictably on the same input, resulting in unreliability in practical applications.

To address the above-mentioned challenges, we propose a novel LLM-based image-to-vector pipeline that robustly converts images to vector graphics, including those generated by text-to-image models. 
Our output vector representation is based on a polygonal scaffold with per-vertex radii, providing a compact and canonical form that simplifies model training. 
Instead of the traditionally used colorful raster inputs, we adopt outline-based raster representations to promote better generalization. Color information is then recovered as a post-processing step by extracting colors from the original input image.
During training, we employ a novel synthetic distortion strategy to simulate variations encountered at test time, which further enhances the model’s robustness to generated images. 
During testing, we introduce a test-time scaling strategy that improves the reliability and consistency of the image-to-vector conversion.


\paragraph{Contributions.}

This paper makes several contributions.
First, we identify the limited generalization ability of existing VLM-based image vectorization methods and propose a new model that performs robustly in real-world vectorization tasks.
Second, we introduce a rounded polygon representation that offers improved geometric accuracy and control-point precision compared to traditional SVG-based formats.
Alongside this, we present an outline-based raster representation that remains robust under varying visual characteristics, such as different anti-aliasing methods.
Third, we incorporate a novel degradation model during training and apply test-time scaling during inference, which significantly enhance the model’s reliability on imperfect or degraded inputs.
Finally, we demonstrate that VectorArk achieves substantially higher fidelity and precision than state-of-the-art methods on both synthetic and real-world datasets.

\section{Related Work}



\paragraph{Classical vectorization and differentiable rendering.}
Classical vectorization methods identify region boundaries by detecting color discontinuities and edge transitions~\cite{Potrace,ch25,xu2022live}. While computationally efficient, they can be sensitive to anti-aliasing, compression artifacts, and noise. This brittleness is particularly problematic for real-world images such as outputs from text-to-image models or low-resolution graphics where artifacts and distortions cause fragmented, topologically inconsistent boundaries. Moreover, lacking semantic understanding, these methods treat all color transitions equally, resulting in over-segmentation and redundant paths.

Differentiable rasterization frameworks like DiffVG~\cite{Li_DiffVG_2020} enable gradient-based optimization of vector parameters, but can produce unstable control points when optimizing for pixel-level reconstruction without geometric priors. Curve regularization techniques~\cite{Yang2016EffectiveClipart} still rely on direct edge tracing and lack semantic structure. In contrast, learning-based approaches such as ours encode strong geometric priors from artist-designed data, enabling smooth, semantically coherent primitives that align with human expectations for vector graphic structure.

\paragraph{Neural SVG generation with multimodal LMs.}
Recent approaches leverage large multimodal models for SVG generation. StarVector~\cite{Rodriguez_StarVector_2025} introduced semantic primitives, and OmniSVG~\cite{yang2025omnisvg} discretized coordinates and commands into tokens; and LLM4SVG~\cite{xing2024llm4svg} integrated SVG tokens into large language models with modular decoders. However, these systems operate within the traditional path-command vocabulary, producing long sequences that amplify small coordinate errors. We demonstrate that a simpler path representation mitigates such error accumulation. Our design is motivated by the notion of reduced degrees of freedom. Fixing endpoints, a cubic Bézier curve requires four parameters, whereas our representation needs only three. Moreover, our formulation enforces piecewise-constant curvature by construction, yielding smoother and more aesthetically pleasing curves.
While recent work on reinforcement learning from rendering feedback~\cite{rodriguez2025renderingawarereinforcementlearningvector} has explored improving SVG generation reliability, these methods do not fully eliminate the unpredictable behavior, and effective strategies for ensuring consistent, high-quality vectorization remain an important challenge.


\section{Method}
\paragraph{Overview.} The input to our model is a raster image, which may be generated by text-to-image models. Our approach aims to automatically convert the input raster into vector graphics while correcting distortions and imperfections present in the original image.
The model is finetuned from a pretrained multimodal large language model (Section \ref{sec:seqmodel}). It takes a raster image as input and autoregressively generates a tokenized vector representation.
The output vector representation (Section \ref{sec:repr}) is constructed as a scaffold of polygons, where each vertex is assigned a radius that defines an inscribed arc at that polygon vertex.
Our input raster employs an outline-based representation (Section \ref{sec:repr}) that is designed to remain robust to discrepancies in visual characteristics between training and testing data.
To further enhance robustness in practical image-to-vector tasks, we introduce a degradation model (Section~\ref{sec:degradation}) during training to simulate visual degradations that may occur during testing, and apply test-time scaling (Section~\ref{sec:TTS}) to improve the reliability of the pipeline.

%


\subsection{Canonical Representations} \label{sec:repr}

\paragraph{Rounded Polygon Representation.}

\begin{figure}[h]
\centering
\setlength{\tabcolsep}{2pt}
\renewcommand{\arraystretch}{0.75}
\begin{tabular}{@{}c@{\hspace{1pt}}c@{\hspace{1pt}}c@{}}
\small (a) Source SVG & \small (b) Line-arc decomposition & \small (c) Polygon vertices \\[2pt]
\includegraphics[width=0.20\columnwidth]{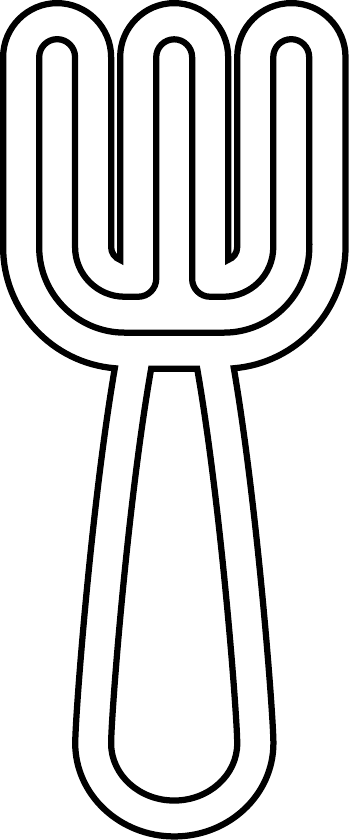} &
\includegraphics[width=0.20\columnwidth]{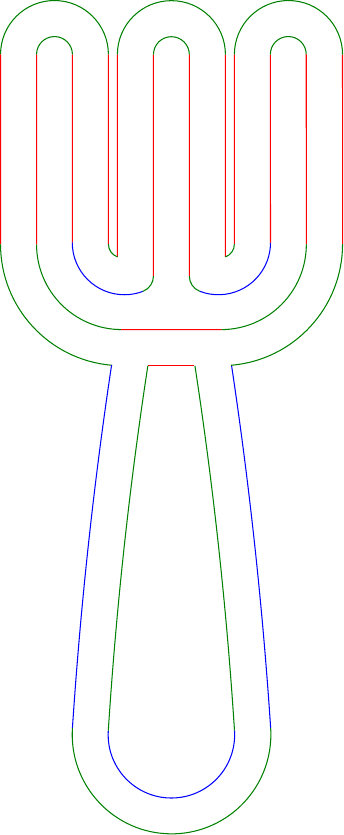} &
\includegraphics[width=0.20\columnwidth]{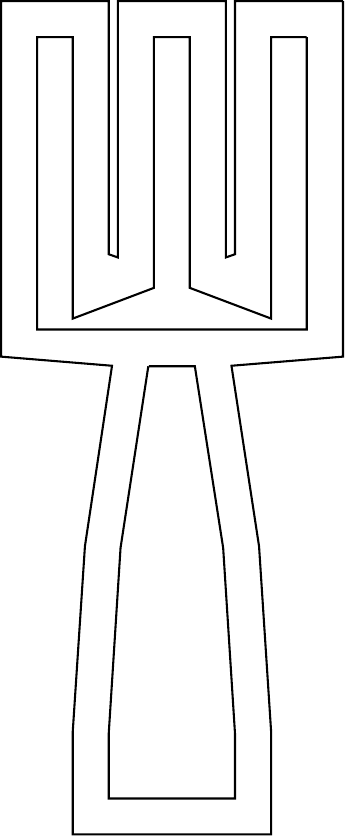}
\end{tabular}
\vspace{-2mm}
\caption{\textbf{Rounded polygon representation example.} Conversion process from an input SVG to our rounded polygon format. (a) Original source SVG with smooth curves. (b) Line-arc decomposition where red segments represent lines and green/blue segments represent circular arcs fitted to the original curves. (c) Corresponding polygon vertices (intersection points) before applying corner radii; the final representation includes these vertices plus the encoded radius.}
\label{fig:representation_example}
\end{figure}

SVGs comprise a wide variety of primitive types—including B\'ezier curves, circles, ellipses, rectangles, polylines, and more—to represent vector graphics. The heterogeneity poses challenges for training generative models. To simplify the learning process, we adopt a canonical and compact representation that expresses SVGs using rounded polygons.
More specifically, we convert each SVG path into a rounded polygon representation independently. 
First, we sample \emph{equidistant} points along the path. 
Then, we fit line and arc primitives to the sampled points while preserving $G^1$ continuity, following the \emph{Cornocopia fitting} algorithm~\cite{Baran_Cornocopia_2010}. After this step, the SVGs are represented using only two primitives: lines and arcs (see Figure~\ref{fig:representation_example}b). As shown in our experiments, this conversion yields a more compact representation than the raw SVG representation and is nearly lossless. 

\begin{figure}[t]
\centering
\includegraphics[width=0.5\columnwidth]{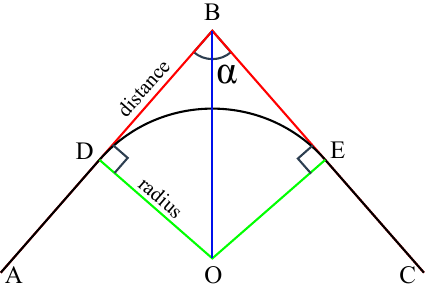}
\caption{\textbf{Rounded polygon vertex computation.} Given fitted line-arc primitives, we compute polygon vertices by extending tangents at arc endpoints to their intersection point.}
\label{fig:arc_representation}
\end{figure}

In the following step, a rounded polygon representation is derived from the lines and arcs.
In this representation, a path is defined as a sequence of triples $\{(x_i, y_i, d_i)\}$, where $(x_i, y_i)$ are the coordinates of the polygon vertices and $d_i$ specifies the roundness at each vertex.
For each line primitive, the two endpoints naturally serve as polygon vertices. For each arc primitive, such as the arc $\wideparen{DE}$ in Figure~\ref{fig:arc_representation}, the tangent vectors at its endpoints, $\overrightarrow{DB}$ and $\overrightarrow{EB}$, intersect at point $B$. The points $D$, $B$, and $E$ (in order) are taken as the polygon vertices. To simplify the representation, any polygon vertex whose two incident edges are collinear is removed, thereby eliminating redundancy without altering the polygon’s shape. 
In addition to the coordinates $(x_i, y_i)$, a roundness value $d_i$ is computed for each polygon vertex.
For vertices such as $B$, formed by the intersection of arc tangents, the roundness value $d_i$ is defined as the distance $BD$ (see Figure~\ref{fig:arc_representation}). This distance-based parameterization enables more stable quantization than using the arc radius directly, as the radius can become excessively large for nearly flat arcs.
The radius of the arc can be computed as $r_i = d_i \tan(\alpha_i/2)$, where $\alpha_i$ is the interior angle $\angle DBE$ at the vertex $B$. 
In this way, the roundness value $d_i$ directly determines the shape of the arc. For other vertices that are not formed by the intersection of arc tangents, the roundness value $d_i$ is set to $-1$, which serves as a special flag indicating that the vertex is an endpoint of a line or an arc.

We handle important edge cases of arcs as follows.
Arcs with negligible curvature are converted to line segments. Large-angle arcs can cause the tangents at their endpoints to intersect far from the arc or not at all; semicircular arcs are a notable example, as their tangents are parallel and never meet.
To address this, arcs are uniformly subdivided such that each resulting sub-arc spans an angle of less than $120^\circ$.

The proposed vector representation employs a rounded polygon format that is both canonical and compact, which simplifies model training, as demonstrated in our experiments.

\paragraph{Outline Raster Representation.}
In image-to-vector conversion tasks, the colors of input rasters can easily exhibit a mismatch between training and testing data. For example, outputs from text-to-image models~\cite{gemini2025gemini2_5, openai2024gpt4o} may have visual characteristics distinct from rendered SVGs, which can hinder the model’s generalization in practical use cases.
To address this issue, we provide the multimodal LLM with a ``decolorized'' raster image as input.
More specifically, we first apply a classic vectorization tool\rebuttaldel{ (Adobe Illustrator Image Trace~\cite{adobeillustrator2025})}\rebuttal{\footnote{We use Adobe Illustrator Image Trace~\cite{adobeillustrator2025} in our experiments and observe comparable behavior with VTracer~\cite{vtracer2020}}} to convert the input raster image into a vector graphic. We then discard its color information and render only the vector paths as a black-and-white raster outline with fixed stroke-width (see Figure~\ref{fig:pipeline}). Although traditional vectorization tools may produce suboptimal vector geometries, they can still faithfully reconstruct the original image. Since our method only requires the rendered output of the vectorization, this step effectively ``decolorizes'' the input image while preserving its geometric structure.
This procedure can be interpreted as a normalization step that maps each input image to a canonical view, thereby improving the model’s robustness to out-of-distribution test data.

Our multimodal LLM is trained to predict colorless outputs from colorless inputs, which greatly simplifies the training process and enhances generalization. During inference, we apply a post-processing step based on a path color extraction method to recover the colors and z-order of output vector paths from the original input image (see the supplementary material for details).
As demonstrated in our experiments, the outline raster representation yields better results than the color raster input, and the color extraction method effectively propagates the input colors to the output vector graphics.


\subsection{Sequence Modeling with Multimodal LLMs} 
\label{sec:seqmodel}

We encode our rounded polygon representation as a text sequence.
Our model is finetuned from a pretrained multimodal LLM (InternVL \cite{chen2024internvl} in our experiments) that takes an image as input and generates the text tokens in an autoregressive manner.
More specifically, we normalize all vector graphics to a $128 \times 128$ viewBox with a 4 unit padding on each side. 
The coordinates $(x_i, y_i)$ and roundness parameters $d_i$ of the target rounded polygon representation are quantized to two decimal places. The sequence ${(x_i, y_i, d_i)}$ is then serialized into text, where each triplet component $x_i, y_i, d_i$ is separated by \texttt{[,]}, consecutive polygon vertices are separated by \texttt{[space]}, and different paths are separated by \texttt{[\textbackslash n]}. We reuse the tokenizer vocabulary from the pretrained multimodal LLM backbone, as this leads to more stable and efficient convergence compared to introducing new tokens. 
Following InternVL \cite{chen2024internvl}, the input image resolution is set to $448 \times 448$. The image is processed by the ViT encoder to produce visual tokens, which are then used as conditioning inputs to the LLM backbone.

During finetuning, we update all parameters of the multimodal LLM, including those of the ViT encoder. We observe that finetuning only the LLM backbone or applying LoRA \cite{hu2021lora} leads to inferior performance.
We adopt the \emph{next-token prediction} objective, using cross-entropy loss over all text tokens to supervise training.

\subsection{Degradation Model}
\label{sec:degradation}

Our model is trained on a dataset of high-quality vector graphics. A straightforward way to prepare training data is to render each vector graphic into a high-fidelity raster image and train the model to reconstruct the original vector from this raster input. However, this approach performs poorly when the input rasters contain degraded or distorted shapes and geometry—such as those produced by text-to-image models. In such cases, the model tends to faithfully reproduce the imperfections of the input raster, leading to low-quality vector graphics.
Another approach is to introduce random noise to the control points of the outline representation described in Section \ref{sec:seqmodel} before rendering it into raster inputs. However, we found that this method does not generalize well to practical image-to-vector tasks.

We propose a vectorization-based degradation model that effectively simulates the types of imperfections commonly observed in practical scenarios.
Specifically, for each training vector graphics, we render it at a randomly sampled reduced resolution (ranging from $224 \times 224$ to $336\times 336$ in our experiments) and apply a random Gaussian blur.
Since classical vectorizers are sensitive to raster resolution, lower-resolution inputs tend to produce poorer vectorization results. 
We leverage this property by applying a classical vectorizer to obtain an imperfect outline, which is then rendered back into a raster image and used as the model’s input during training.
As demonstrated in our experiments, the proposed degradation model enables our method to generalize effectively across diverse image inputs, including those generated by text-to-image models.


\subsection{Test-Time Scaling}
\label{sec:TTS}

Compared to classical vectorizers, LLM-based image-to-vector methods can succeed or fail unpredictably on the same input when evaluated with different random seeds. To address this issue, we apply test-time scaling during inference. For each input image, we generate $N$ candidate vector graphics through stochastic decoding of our image-to-vector model. We then recover the colors and z-order of the output vector paths from the input raster. Each candidate vector graphic is rendered back into a raster image, and both these rendered images and the original input raster are encoded into embeddings using the frozen DINO-ViT-B/16 encoder \cite{Caron_DINO_2021}. The final output is selected as the candidate whose rendered raster has the highest cosine similarity with the input raster in the embedding space. Among various feature extractors we tested, DINO yielded the most reliable results, consistent with observations from StarVector~\cite{Rodriguez_StarVector_2025}, significantly improving the robustness of our pipeline, which is crucial for practical image-to-vector applications.

%% file: sec/3_finalcopy.tex
\section{Experiments}

\input{sec/exp}

\section{Conclusion}
\rebuttaldel{We introduced a geometry-first VLM pipeline for raster-to-SVG vectorization, combining a novel rounded polygon representation, a new degradation model, and a test-time scaling strategy. 
Across benchmarks and difficult real-world inputs, the method yields compact, faithful, and robust SVGs. 
Ablations clarify why each component matters, offering practical guidance for vector graphics generative models. 
Our representation targets \emph{medium-complexity} graphics; highly detailed illustrations and text/gradient effects are simplified.
Extremely out-of-distribution inputs may benefit from larger backbones or modest adaptation.
Future work includes scaling to richer primitives under controlled constraints and integrating a learned appearance module while preserving geometric fidelity.}
\rebuttal{We introduced a geometry-first VLM pipeline for raster-to-SVG vectorization, combining a novel rounded polygon representation, a new degradation model, and a test-time scaling strategy. Across benchmarks and difficult real-world inputs, the method yields compact, faithful, and robust SVGs. Ablations clarify why each component matters, offering practical guidance for vector graphics generative models. Our representation targets \emph{medium-complexity} graphics; highly detailed illustrations and text/gradient effects are simplified, and dense local path structure remains challenging (see supplementary). Future work includes scaling to richer primitives under controlled constraints and integrating a learned appearance module while preserving geometric fidelity.}

%% file: sec/exp.tex
\subsection{Experimental Setup}

\paragraph{Implementation Details.}
Our model is finetuned from InternVL2-1B \cite{chen2024internvl}, \rebuttaldel{updating all parameters including the base LLM and ViT encoder}\rebuttal{ in an end-to-end manner}. The ViT encoder processes \rebuttaldel{outline-based raster input} \rebuttal{outline rasters} at $448\times 448$ resolution. We employ AdamW optimizer \cite{loshchilov2019adamw} with cosine learning rate decay, initial learning rate $10^{-4}$, batch size $256$, and $250K$ iterations. Training uses $\sim 5M$ SVGs \rebuttaldel{including} \rebuttal{from} icons, logos, and flat graphics, with random rotation and scaling augmentation. The degradation model is \rebuttaldel{randomly} dropped with probability $25\%$ to ensure robustness on clean inputs.

During training, we normalize the ground-truth SVGs to a $128 \times 128$ viewBox and apply a classic vectorizer to create stroke-only outlines. 
During inference, given a raster input, we apply the \emph{same} vectorization pipeline and parameters used during training to extract a stroke-only outline. This acts as a ``normalization'' step to avoid training-testing mismatch.

\paragraph{Benchmark Datasets.} 
We benchmark our method on the two latest SVG generation benchmarks: \textbf{SArena}~\cite{wang2025internsvg} and \textbf{SVGenius}~\cite{chen2025svgeniusbenchmarkingllmssvg}. To facilitate systematic evaluation across varying levels of geometric complexity, we stratify samples into difficulty tiers following SVGenius. For SArena, we apply a primitive-count-based categorization (\emph{Easy}: $<64$, \emph{Medium}: $64$--$128$, \emph{Hard}: $>128$ path primitives), while for SVGenius, we adopt their predefined complexity levels. 

\paragraph{Evaluation Metrics.} 
Following OmniSVG \cite{yang2025omnisvg}, we assess reconstruction quality using four metrics: \textbf{LPIPS}~\cite{Zhang_LPIPS_2018}, \textbf{SSIM}~\cite{Wang_SSIM_2004}, \textbf{DINO}~\cite{Caron_DINO_2021} (ViT-B/16 cosine similarity), and \textbf{MSE}. These metrics jointly capture perceptual, structural, and pixel-wise fidelity of generated SVGs relative to ground truth. 

\begin{table*}[t]
\centering
\caption{Quantitative comparison on SArena and SVGenius benchmarks across difficulty levels. Our method achieves the best performance across all metrics and complexity categories. Bold indicates best score for each metric.}
\label{tab:comp_prior}
\begin{minipage}{0.49\textwidth}
\centering
\resizebox{\textwidth}{!}{
\begin{tabular}{@{}l|ccccc@{}}
\toprule
\multicolumn{6}{c}{\textbf{SArena}} \\
\midrule
\textbf{Diff./Metric} & \textbf{GPT-4o} & \textbf{Gemini} & \textbf{OmniSVG} & \textbf{StarVector} & \textbf{Ours} \\
\midrule
E / SSIM $\uparrow$ & 0.681 & 0.622 & 0.823 & 0.876 & \textbf{0.937} \\
E / LPIPS $\downarrow$ & 0.205 & 0.253 & 0.099 & 0.069 & \textbf{0.031} \\
E / MSE $\downarrow$ & 0.095 & 0.121 & 0.036 & 0.032 & \textbf{0.011} \\
E / Dino $\uparrow$ & 0.972 & 0.944 & 0.98 & 0.969 & \textbf{0.992} \\
\midrule
M / SSIM $\uparrow$ & 0.53 & 0.493 & 0.6 & 0.75 & \textbf{0.895} \\
M / LPIPS $\downarrow$ & 0.284 & 0.323 & 0.251 & 0.142 & \textbf{0.058} \\
M / MSE $\downarrow$ & 0.133 & 0.163 & 0.125 & 0.062 & \textbf{0.013} \\
M / Dino $\uparrow$ & 0.958 & 0.932 & 0.903 & 0.949 & \textbf{0.981} \\
\midrule
H / SSIM $\uparrow$ & 0.47 & 0.441 & 0.518 & 0.626 & \textbf{0.857} \\
H / LPIPS $\downarrow$ & 0.334 & 0.382 & 0.324 & 0.252 & \textbf{0.093} \\
H / MSE $\downarrow$ & 0.151 & 0.169 & 0.123 & 0.101 & \textbf{0.022} \\
H / Dino $\uparrow$ & 0.931 & 0.897 & 0.898 & 0.902 & \textbf{0.975} \\
\bottomrule
\end{tabular}
}
\end{minipage}
\hfill
\begin{minipage}{0.49\textwidth}
\centering
\resizebox{\textwidth}{!}{
\begin{tabular}{@{}l|ccccc@{}}
\toprule
\multicolumn{6}{c}{\textbf{SVGenius}} \\
\midrule
\textbf{Diff./Metric} & \textbf{GPT-4o} & \textbf{Gemini} & \textbf{OmniSVG} & \textbf{StarVector} & \textbf{Ours} \\
\midrule
E / SSIM $\uparrow$ & 0.673 & 0.611 & 0.84 & 0.89 & \textbf{0.944} \\
E / LPIPS $\downarrow$ & 0.19 & 0.244 & 0.07 & 0.046 & \textbf{0.028} \\
E / MSE $\downarrow$ & 0.094 & 0.121 & 0.027 & 0.019 & \textbf{0.008} \\
E / Dino $\uparrow$ & 0.976 & 0.951 & 0.985 & 0.993 & \textbf{0.995} \\
\midrule
M / SSIM $\uparrow$ & 0.572 & 0.539 & 0.674 & 0.71 & \textbf{0.868} \\
M / LPIPS $\downarrow$ & 0.28 & 0.328 & 0.204 & 0.203 & \textbf{0.08} \\
M / MSE $\downarrow$ & 0.087 & 0.097 & 0.058 & 0.058 & \textbf{0.015} \\
M / Dino $\uparrow$ & 0.942 & 0.914 & 0.94 & 0.921 & \textbf{0.977} \\
\midrule
H / SSIM $\uparrow$ & 0.566 & 0.536 & 0.638 & 0.672 & \textbf{0.83} \\
H / LPIPS $\downarrow$ & 0.295 & 0.326 & 0.248 & 0.258 & \textbf{0.12} \\
H / MSE $\downarrow$ & 0.079 & 0.092 & 0.065 & 0.059 & \textbf{0.023} \\
H / Dino $\uparrow$ & 0.928 & 0.918 & 0.918 & 0.893 & \textbf{0.958} \\
\bottomrule
\end{tabular}
}
\end{minipage}
\end{table*}

\subsection{Comparison with Prior Art}
For comparative evaluation, we include recent state-of-the-art multimodal baselines: \textbf{OmniSVG}~\cite{yang2025omnisvg}, \textbf{StarVector}~\cite{Rodriguez_StarVector_2025}, as well as proprietary multimodal models \textbf{GPT-4o}~\cite{openai2024gpt4o} and \textbf{Gemini}~\cite{gemini2025gemini2_5}. 
For OmniSVG and StarVector, we use their models with highest performance. 
For GPT-4o and Gemini, we adopt the prompts from SVGenius~\cite{chen2025svgeniusbenchmarkingllmssvg}. All baselines employ identical test-time scaling configurations to ensure fair comparison.

\paragraph{Quantitative Evaluation.}
Quantitative results are summarized in Table~\ref{tab:comp_prior}. 
Although our model has fewer parameters compared to the baselines,
we observe a consistent and significant improvement across all four metrics, LPIPS, SSIM, DINO, and MSE across both SArena and SVGenius benchmarks, with particularly pronounced gains on higher complexity tiers. 
\rebuttal{This advantage persists even in the single-sample setting without test-time scaling ($N\!=\!1$) on SVGenius, where our method still outperforms the strongest baseline.}


\paragraph{Qualitative Evaluation.}

\begin{figure}[t]
\centering
\footnotesize
\setlength{\tabcolsep}{1.5pt}
\renewcommand{\arraystretch}{0.75}
\resizebox{0.9\columnwidth}{!}{
\begin{tabular}{@{}ccccc@{}}
\scriptsize Input Raster & \scriptsize StarVector 8B & \scriptsize OmniSVG & \scriptsize GPT-4o & \scriptsize \textbf{Ours} \\[1pt]
\includegraphics[width=0.17\columnwidth]{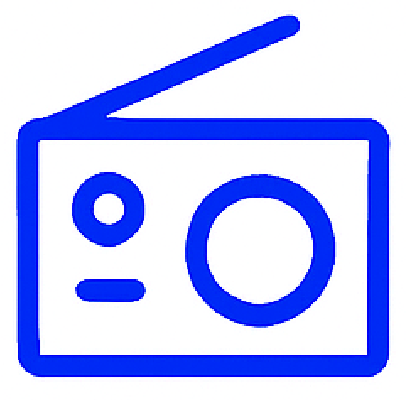} &
\includegraphics[width=0.17\columnwidth]{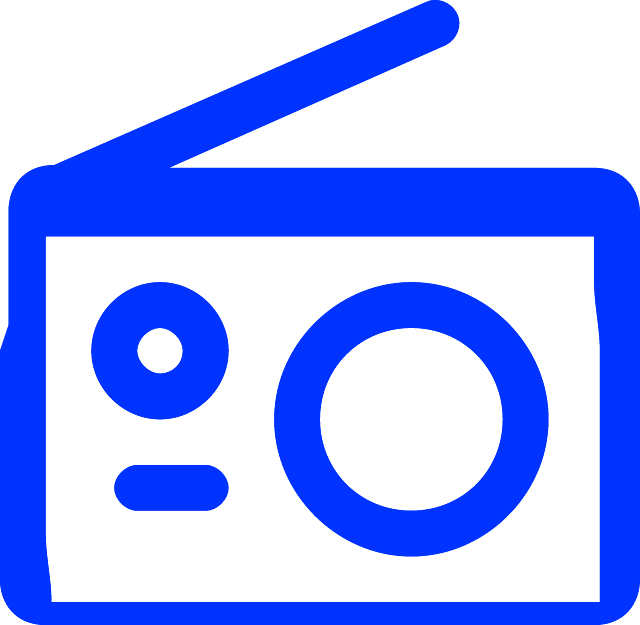} &
\includegraphics[width=0.17\columnwidth]{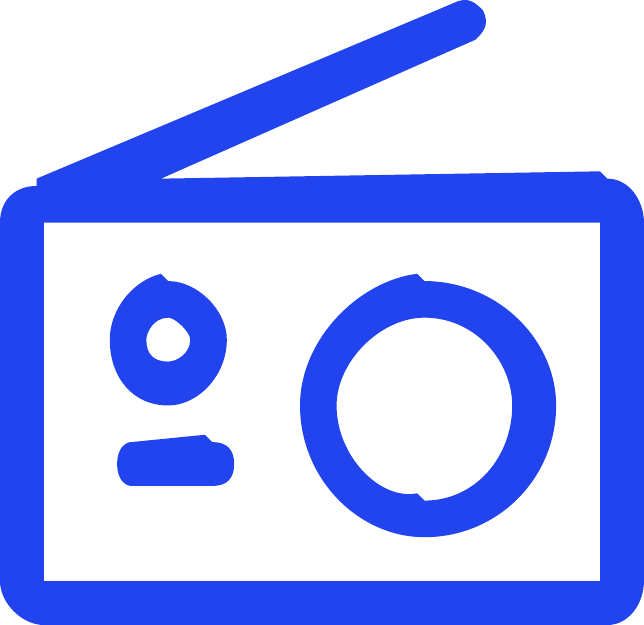} &
\includegraphics[width=0.17
\columnwidth]{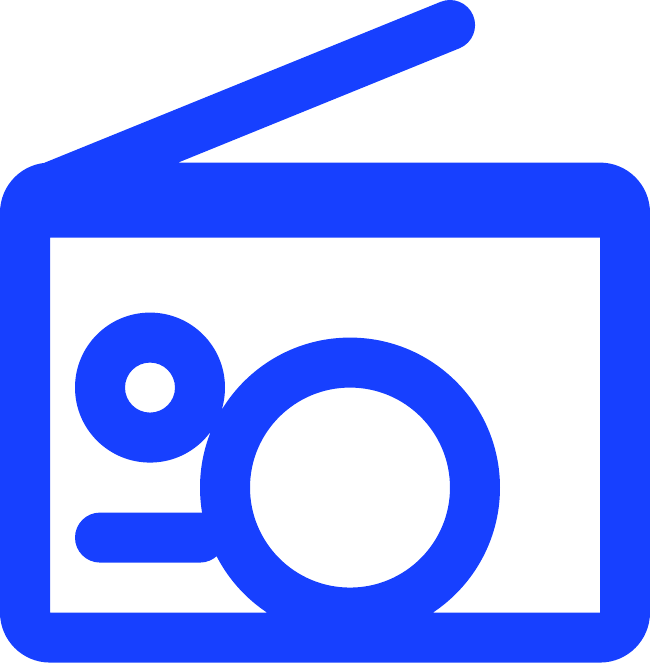} &
\includegraphics[width=0.17\columnwidth]{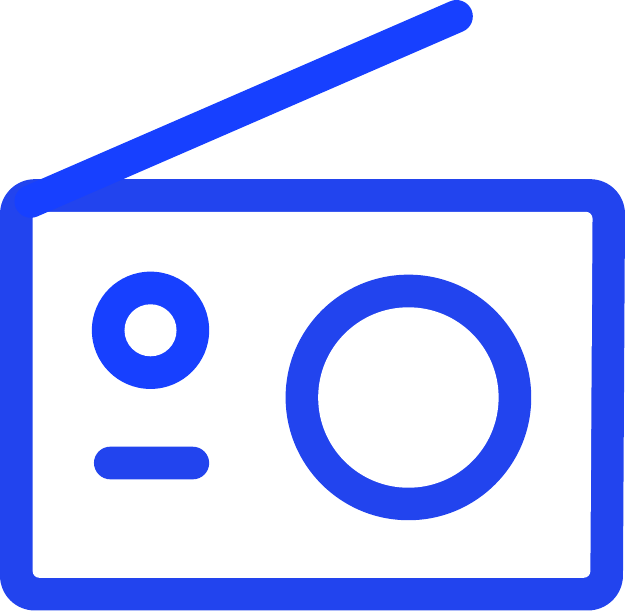} \\[2pt]
\includegraphics[width=0.17\columnwidth]{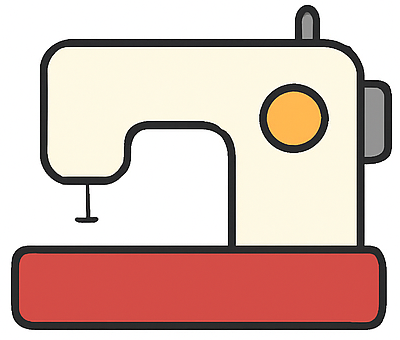} &
\includegraphics[width=0.17\columnwidth]{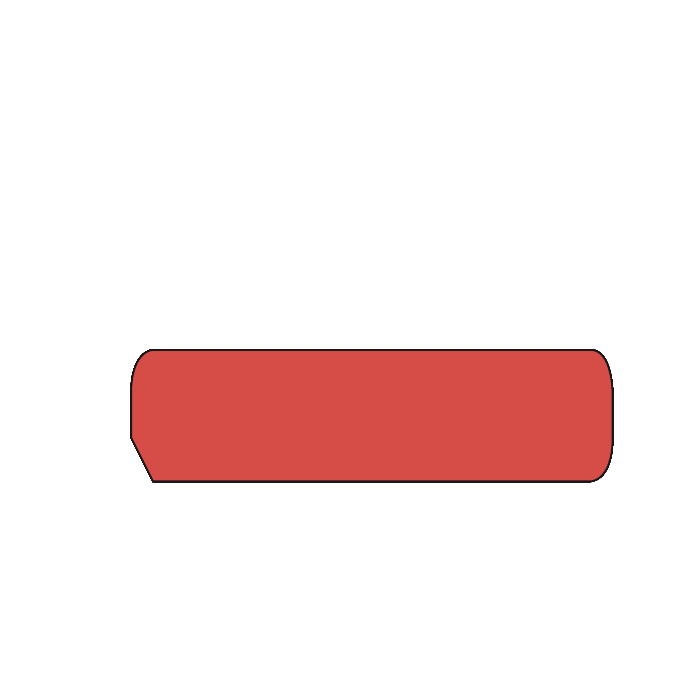} &
\includegraphics[width=0.17\columnwidth]{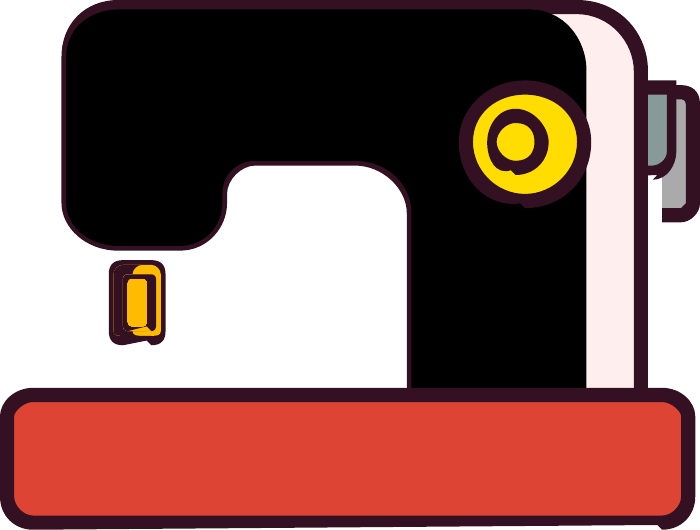} &
\includegraphics[width=0.17
\columnwidth]{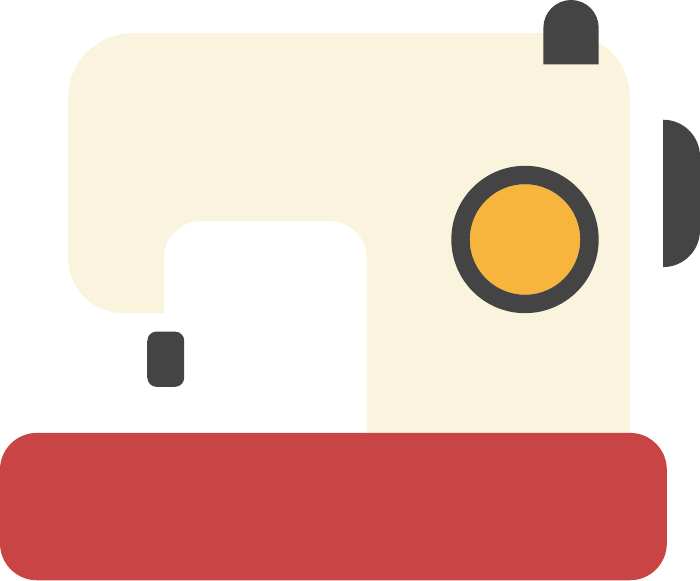} &
\includegraphics[width=0.17\columnwidth]{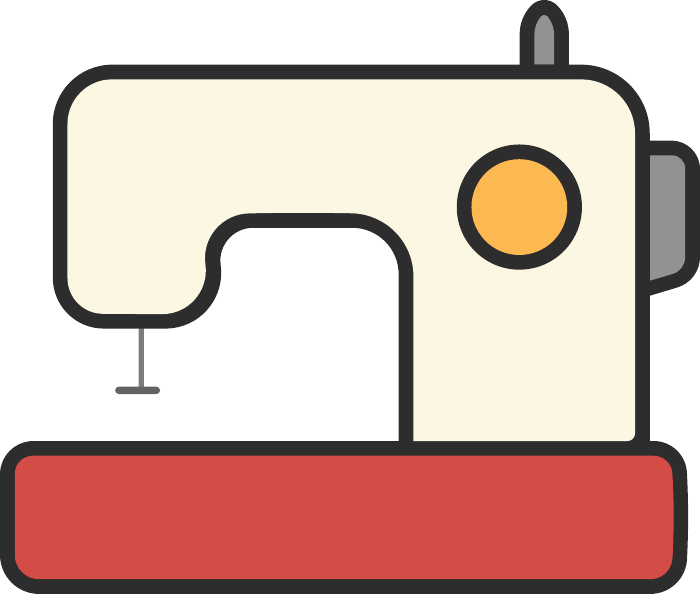} \\[2pt]
\includegraphics[width=0.17\columnwidth]{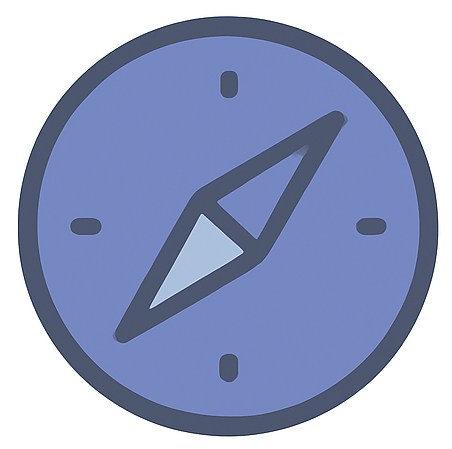} &
\includegraphics[width=0.17\columnwidth]{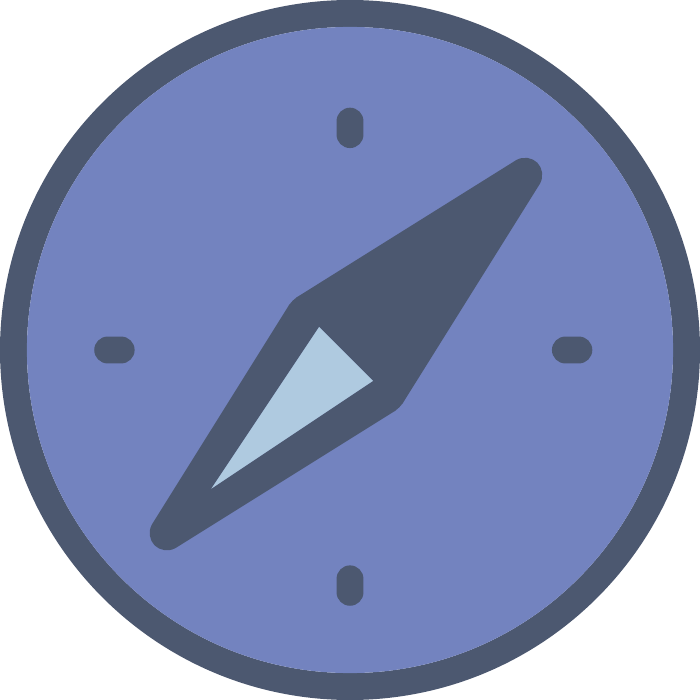} &
\includegraphics[width=0.17\columnwidth]{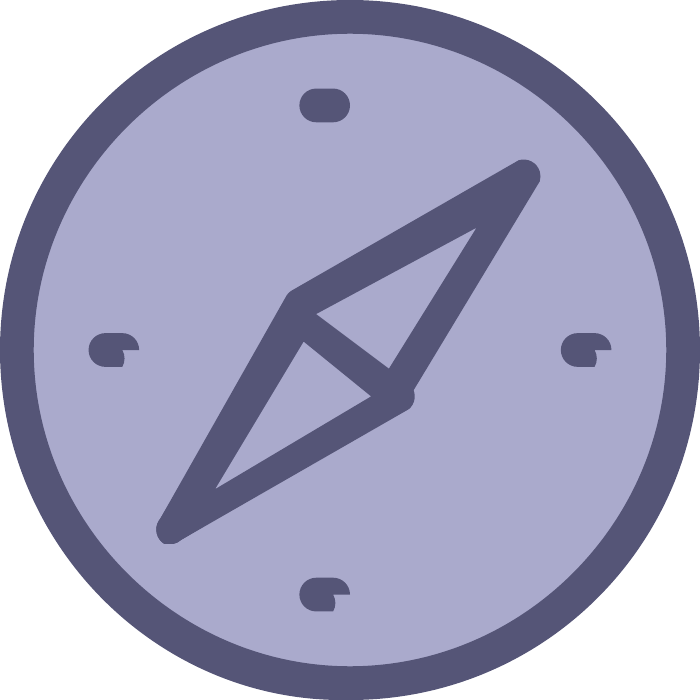} &
\includegraphics[width=0.17
\columnwidth]{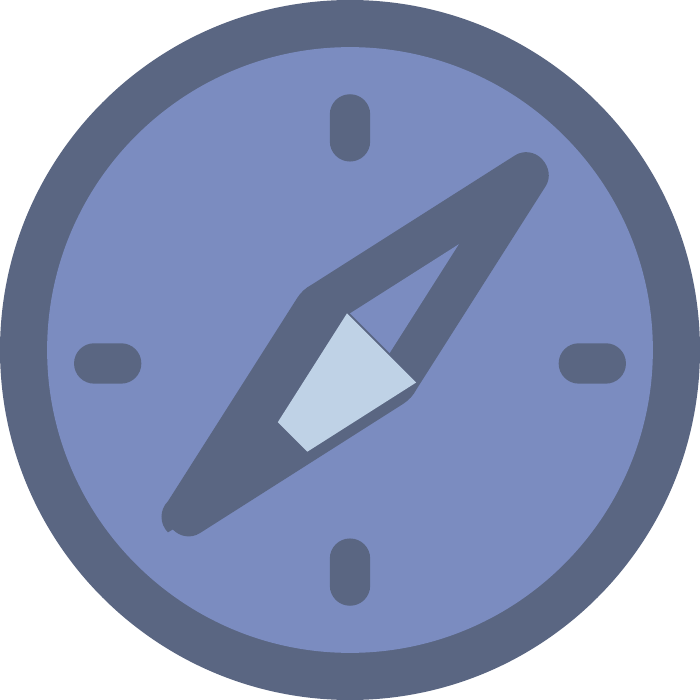} &
\includegraphics[width=0.17\columnwidth]{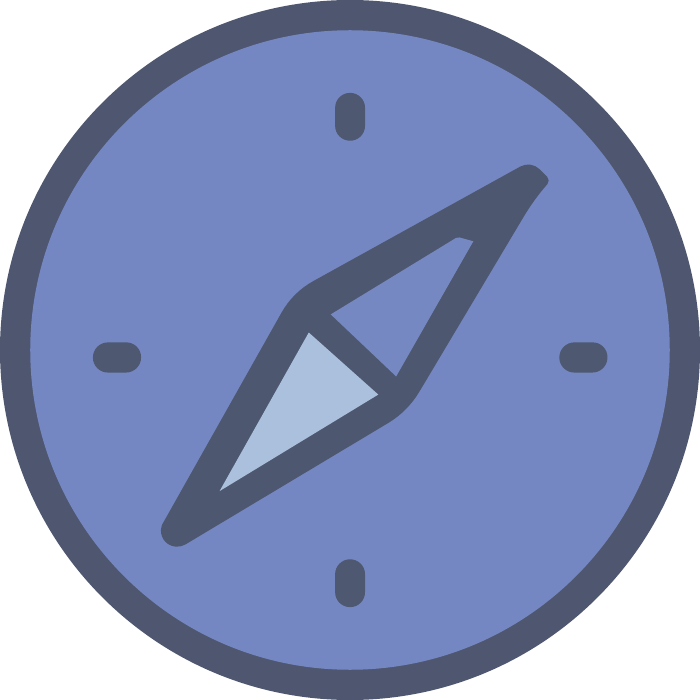} \\[2pt]
\includegraphics[width=0.17\columnwidth]{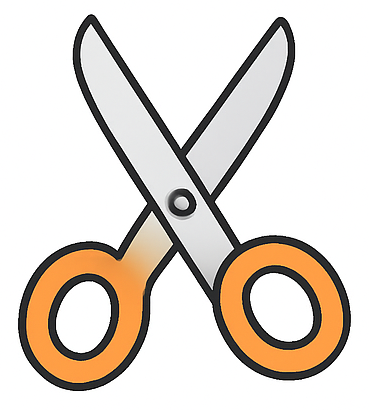} &
\includegraphics[width=0.17\columnwidth]{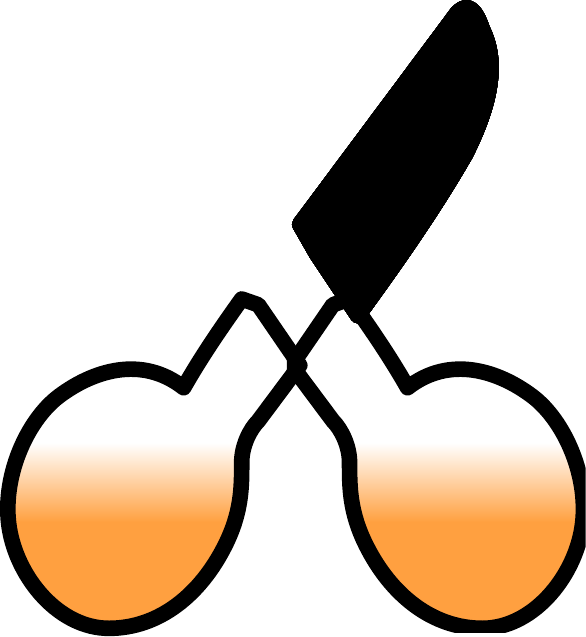} &
\includegraphics[width=0.17\columnwidth]{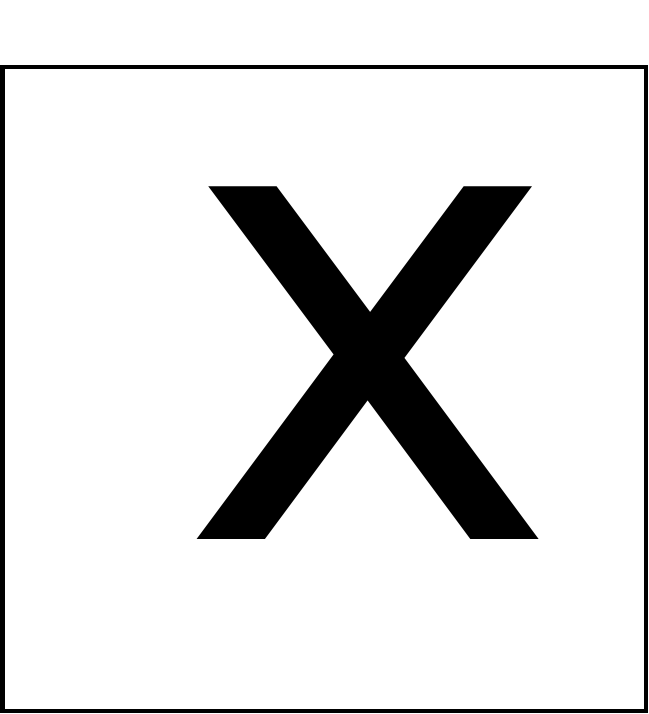} &
\includegraphics[width=0.17\columnwidth]{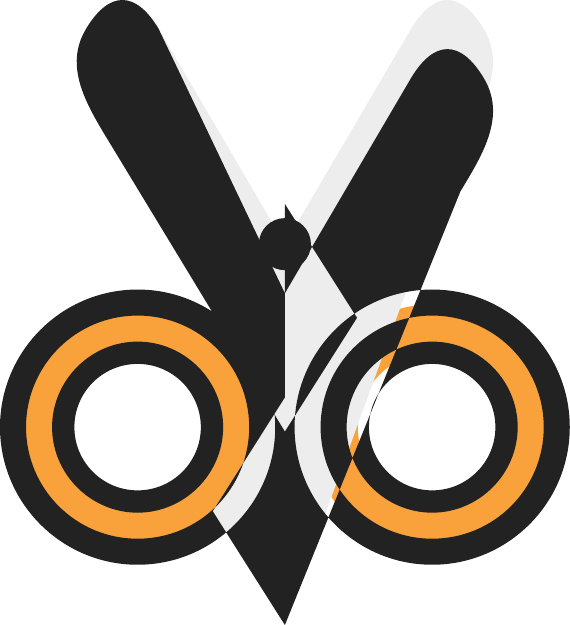} &
\includegraphics[width=0.17\columnwidth]{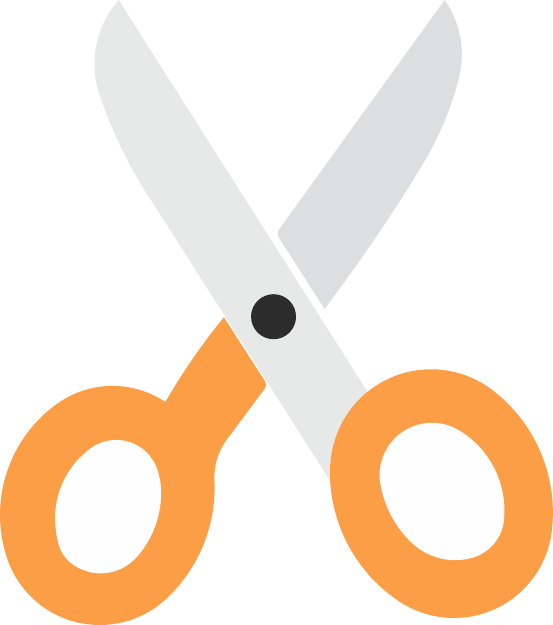} \\[2pt]
\includegraphics[width=0.17\columnwidth]{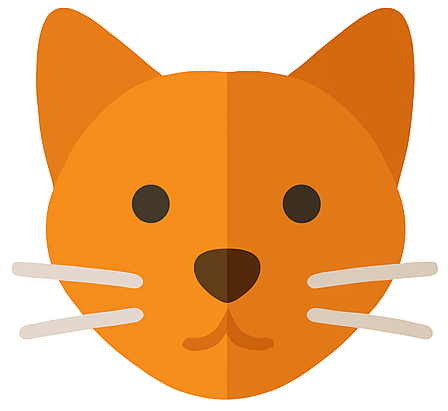} &
\includegraphics[width=0.17\columnwidth]{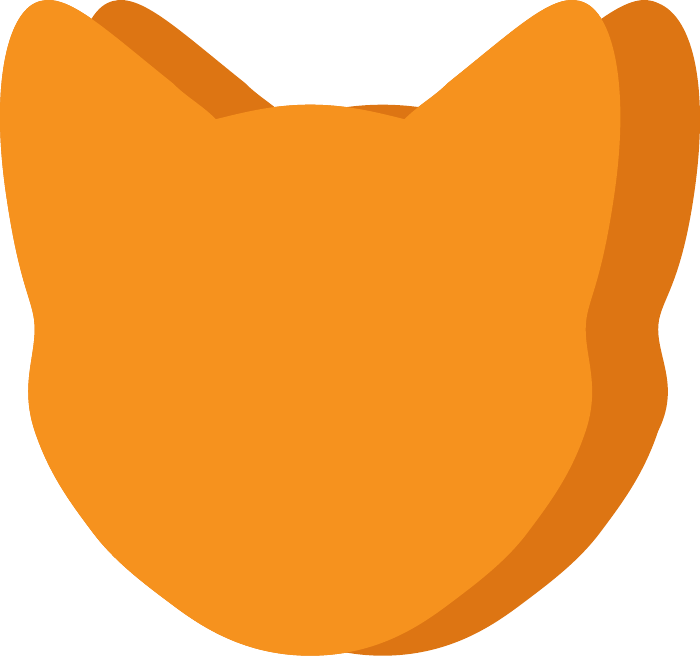} &
\includegraphics[width=0.17\columnwidth]{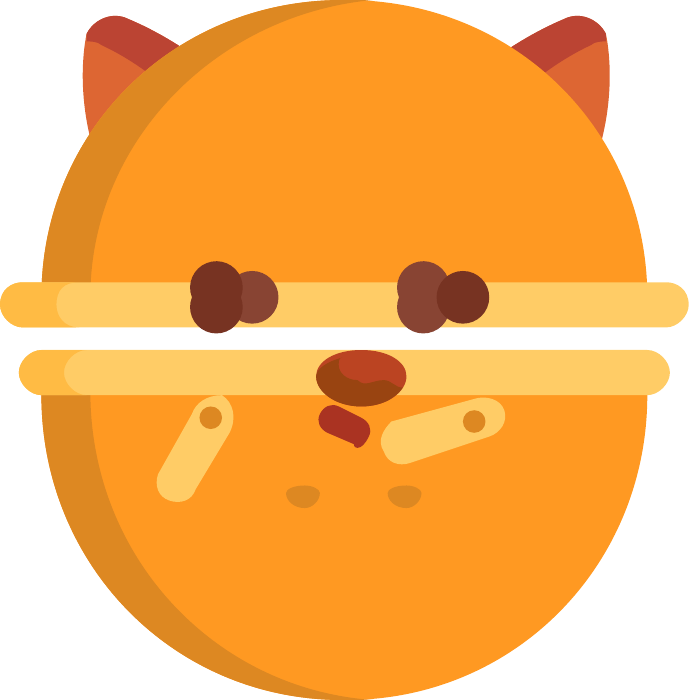} &
\includegraphics[width=0.17\columnwidth]{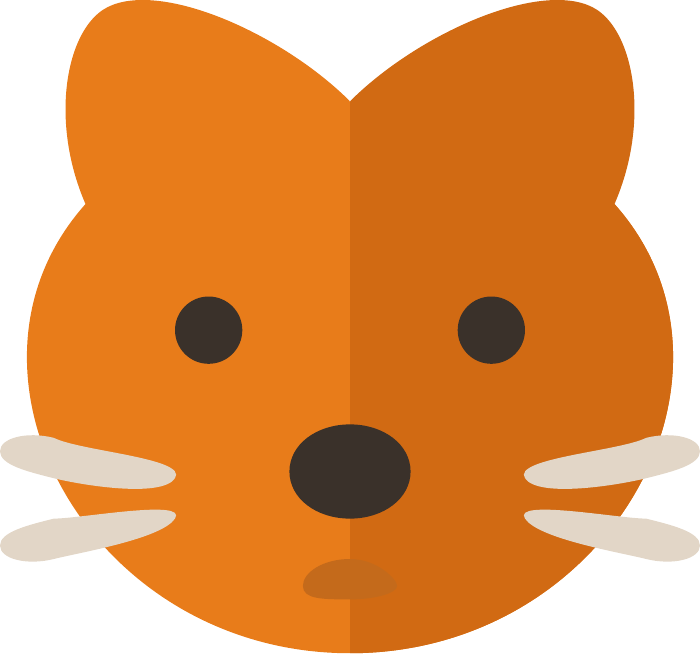} &
\includegraphics[width=0.17\columnwidth]{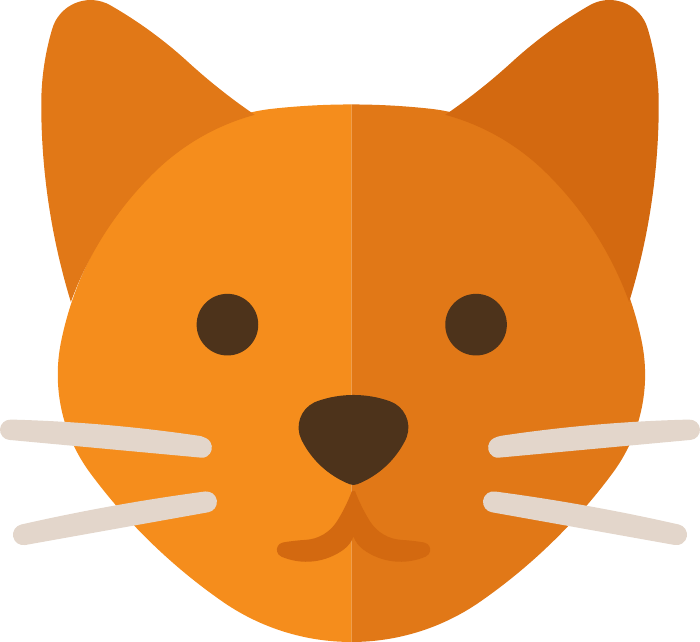} \\[2pt]
\includegraphics[width=0.17\columnwidth]{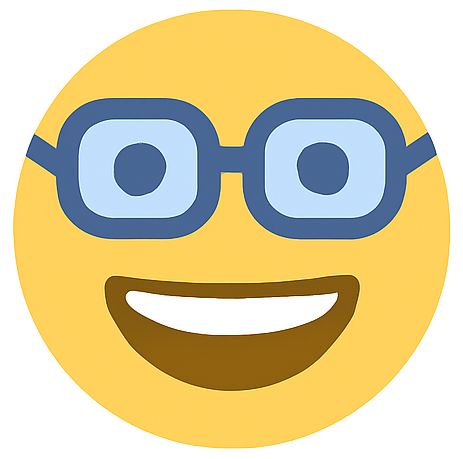} &
\includegraphics[width=0.17\columnwidth]{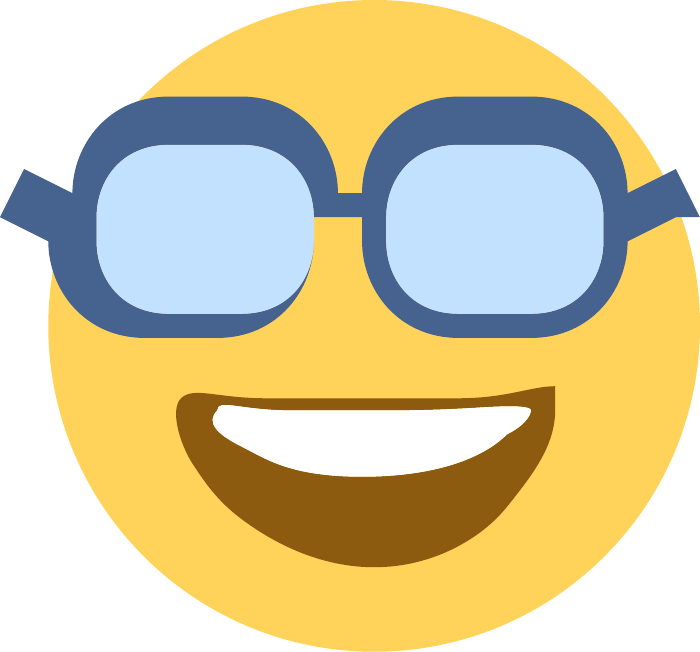} &
\includegraphics[width=0.17\columnwidth]{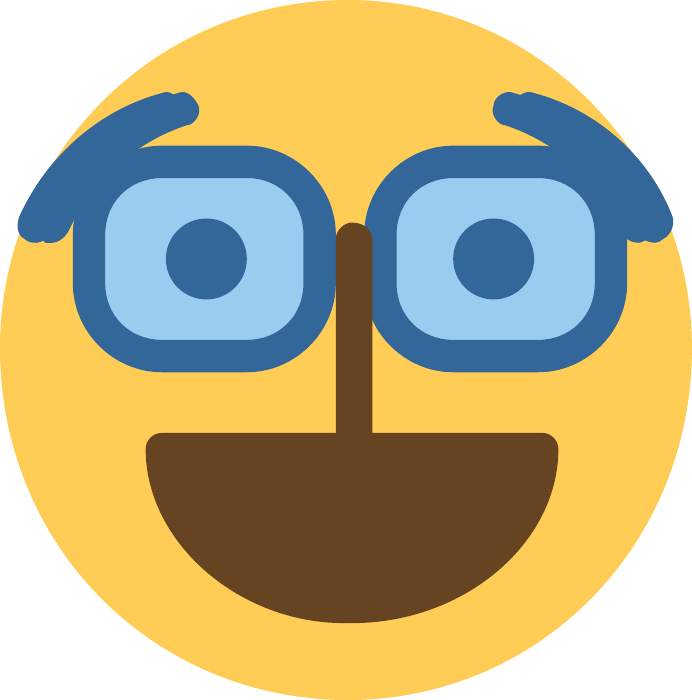} &
\includegraphics[width=0.17\columnwidth]{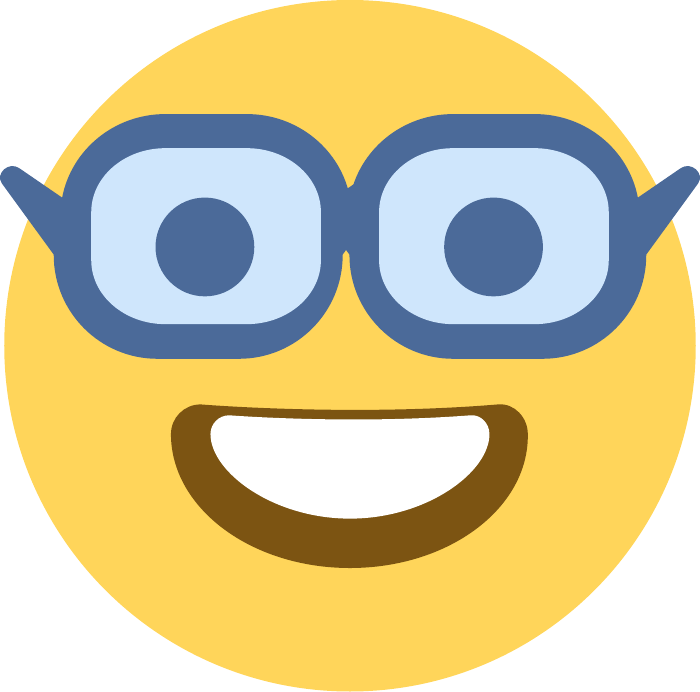} &
\includegraphics[width=0.17\columnwidth]{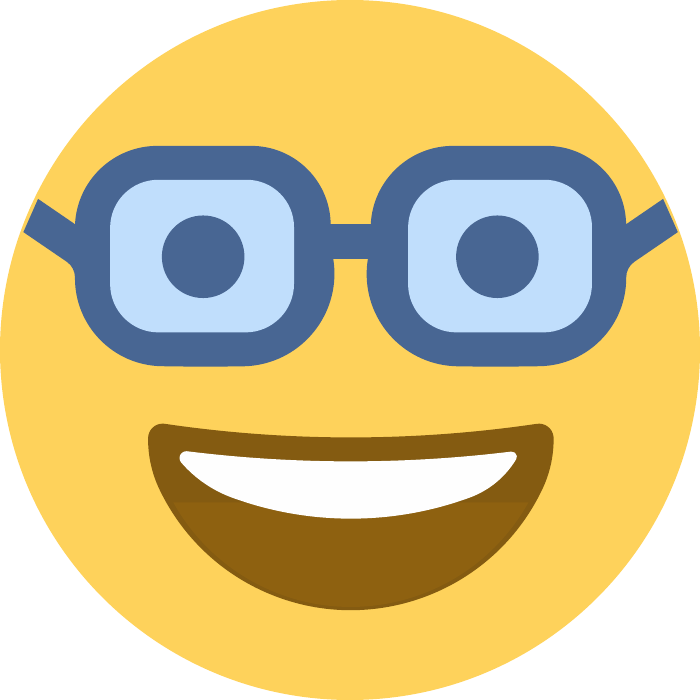} \\[2pt]
\includegraphics[width=0.17\columnwidth]{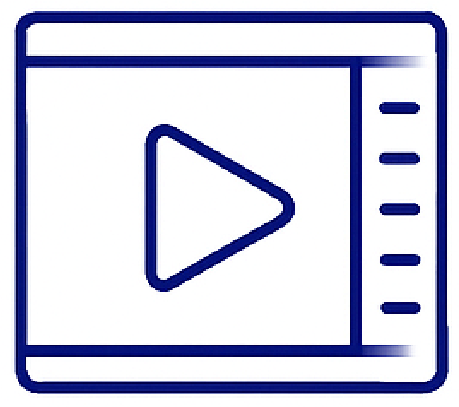} &
\includegraphics[width=0.17\columnwidth]{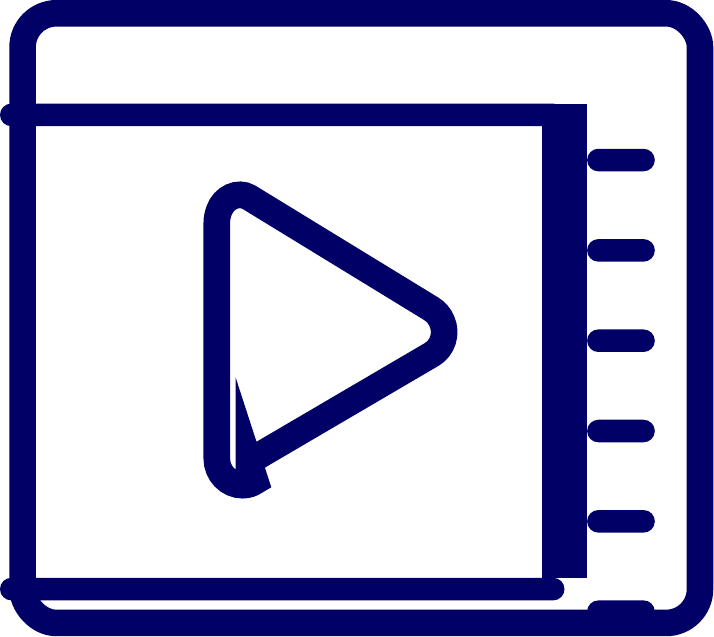} &
\includegraphics[width=0.17\columnwidth]{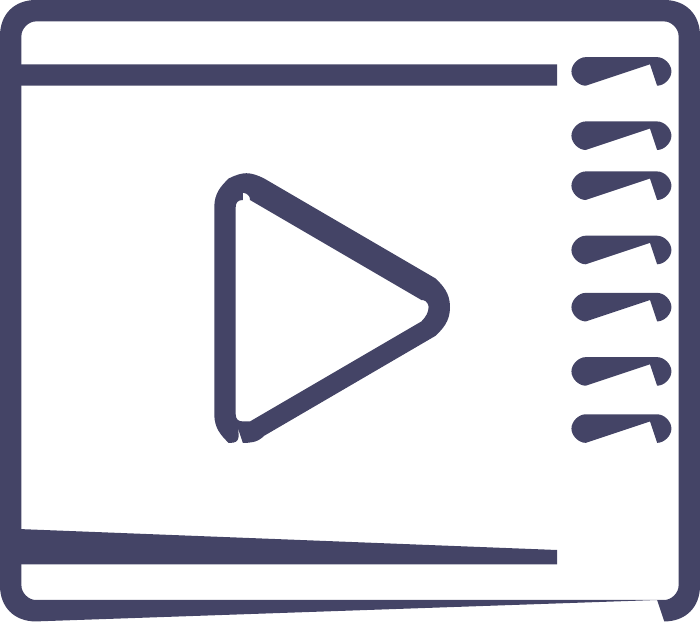} &
\includegraphics[width=0.17\columnwidth]{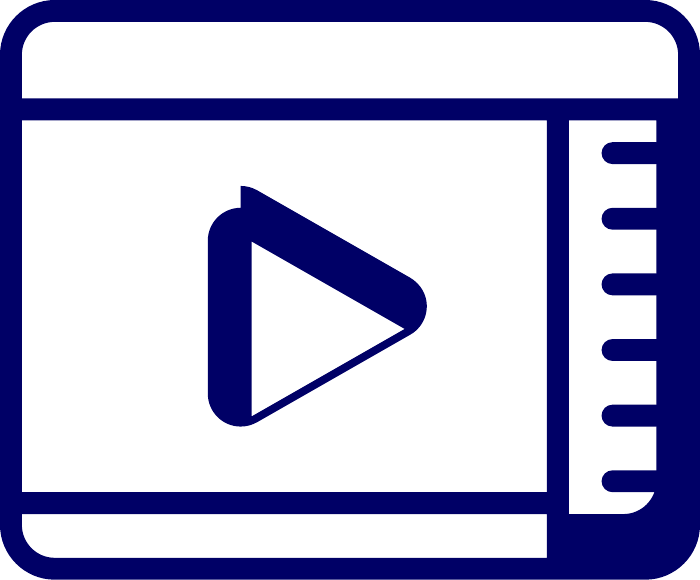} &
\includegraphics[width=0.17\columnwidth]{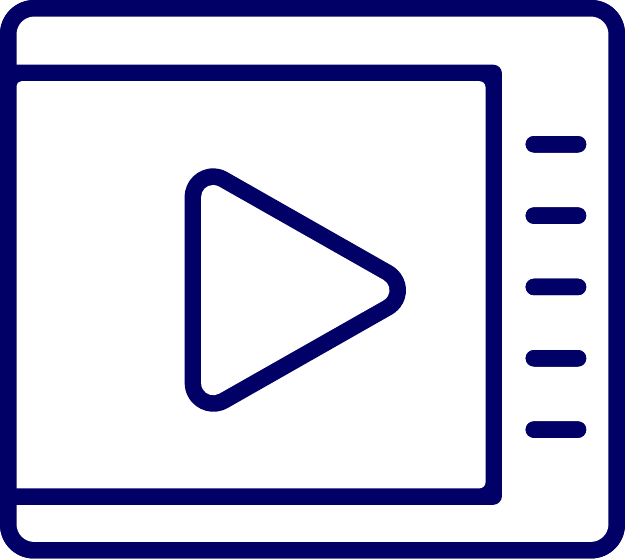} \\
\end{tabular}
}
\vspace{-2mm}
\caption{\textbf{Qualitative comparison on text-to-image generated inputs.}}
\label{fig:qual_comparison}
\end{figure}

Figure~\ref{fig:qual_comparison} presents qualitative comparisons on challenging inputs generated by text-to-image models (GPT-4o~\cite{openai2024gpt4o} and Gemini~\cite{gemini2025gemini2_5}). 
These generated images often contain artifacts and distorted shapes that substantially degrade vectorization quality. 
While all baselines struggle with such imperfect inputs and produce imprecise geometry, our approach demonstrates significantly superior robustness. 

\subsection{Ablation Study}

\paragraph{Ablation on Rounded Polygon Representation}
We analyze the impact of our proposed rounded polygon representation by keeping the dataset, model architecture, and training setup identical, while substituting our representation with OmniSVG's and StarVector's representation. 
For the OmniSVG~\cite{yang2025omnisvg} representation, all SVG commands are restricted to MoveTo (\texttt{M}), LineTo (\texttt{L}), Arc (\texttt{A}), CubicBézier (\texttt{C}), and ClosePath (\texttt{Z}) using only absolute coordinates. 
We enforce a fixed viewBox with standardized padding, and quantize all coordinates to two decimal places of precision following our setup. 
All SVG constructs—regardless of their original type—are expanded into \texttt{<path>} elements, and their corresponding ``\texttt{d}'' attributes are extracted and concatenated into a single string representation, with individual paths separated by newline characters. 
For the StarVector~\cite{Rodriguez_StarVector_2025} representation, we adopt their native SVG format which uses the input raw SVG structure.

\begin{table}[t]
\centering
\caption{Ablation study comparing OmniSVG representation, StarVector representation, and our proposed line-arc hybrid representation on SVGenius dataset across difficulty levels. Bold indicates best score for each metric.}
\label{tab:ablation_path}
\resizebox{\columnwidth}{!}{
\begin{tabular}{@{}ll|ccc@{}}
\toprule
\textbf{Difficulty} & \textbf{Metric} & \textbf{OmniSVG} & \textbf{StarVector} & \textbf{Ours} \\
\midrule
Easy & SSIM $\uparrow$ & \textbf{0.987} & 0.95 & 0.944 \\
Easy & LPIPS $\downarrow$ & \textbf{0.01} & 0.029 & 0.028 \\
Easy & MSE $\downarrow$ & \textbf{0.004} & 0.01 & 0.008 \\
Easy & DinoScore $\uparrow$ & 0.994 & 0.992 & \textbf{0.995} \\
\midrule
Medium & SSIM $\uparrow$ & 0.858 & 0.697 & \textbf{0.868} \\
Medium & LPIPS $\downarrow$ & 0.111 & 0.242 & \textbf{0.08} \\
Medium & MSE $\downarrow$ & 0.025 & 0.058 & \textbf{0.015} \\
Medium & DinoScore $\uparrow$ & 0.959 & 0.908 & \textbf{0.977} \\
\midrule
Hard & SSIM $\uparrow$ & 0.743 & 0.628 & \textbf{0.83} \\
Hard & LPIPS $\downarrow$ & 0.204 & 0.31 & \textbf{0.12} \\
Hard & MSE $\downarrow$ & 0.04 & 0.066 & \textbf{0.023} \\
Hard & DinoScore $\uparrow$ & 0.923 & 0.866 & \textbf{0.958} \\
\bottomrule
\end{tabular}
}
\end{table}








The quantitative evaluation is presented in Table~\ref{tab:ablation_path}.
\rebuttaldel{all the variants achieve comparable quality on simple vector graphics (Easy category), where the OmniSVG representation performs slightly better on pixel-level metrics while our representation marginally leads in semantic similarity (DinoScore). However, our representation substantially outperforms the baselines on medium and hard categories, showcasing the effectiveness of our canonical vector representation.}
\rebuttal{Our rounded-polygon representation substantially outperforms the baselines on the medium and hard splits, highlighting the benefit of the proposed canonical representation. On the Easy split, the raw SVG-command representation attains slightly stronger pixel-level metrics while our representation leads in semantic similarity (DINO); we analyze this behavior in detail in the supplementary material.}

\begin{table}[t]
\centering
\caption{\textbf{Token efficiency comparison across difficulty levels.} Our representation achieves substantial token reductions (27.9--46.6\%) while maintaining high reconstruction quality (all DINO scores $>$ 0.99).}
\resizebox{\columnwidth}{!}{
\begin{tabular}{@{}l|ccc@{}}
\toprule
\textbf{Dataset/Difficulty} & \textbf{Ours} & \textbf{OmniSVG} & \textbf{Savings (\%)} \\
\midrule
\multicolumn{4}{l}{\textit{SArena Icon Generation}} \\
\midrule
Easy & 642 & 890 & 27.9 \\
Medium & 1568 & 2332 & 32.7 \\
Hard & 3130 & 4637 & 32.5 \\
\midrule
\multicolumn{4}{l}{\textit{SVGenius}} \\
\midrule
Easy & 613 & 960 & 36.1 \\
Medium & 2694 & 5046 & 46.6 \\
Hard & 4345 & 7838 & 44.6 \\
\bottomrule
\end{tabular}
}
\label{tab:tokenization_comparison}
\end{table}

\paragraph{Efficiency of the Rounded Polygon Representation.}
\rebuttaldel{Beyond vectorization quality, our representation also provides computational efficiency gains.}
\rebuttal{Our representation is also more token-efficient.}
As shown in Table~\ref{tab:tokenization_comparison}, we achieve consistent token reductions ranging from 27.9\% to 46.6\% across all difficulty levels on both benchmarks while maintaining nearly lossless reconstruction quality (DINO score $>$ 0.99). \rebuttaldel{These efficiency gains translate directly to reduced memory footprint and faster inference time.}
\rebuttal{This directly reduces memory footprint and inference cost. In practice, single-SVG generation takes only $3$--$4$\,s on an A100.}

\begin{figure*}[!htb]
\centering
\includegraphics[width=1.0\linewidth]{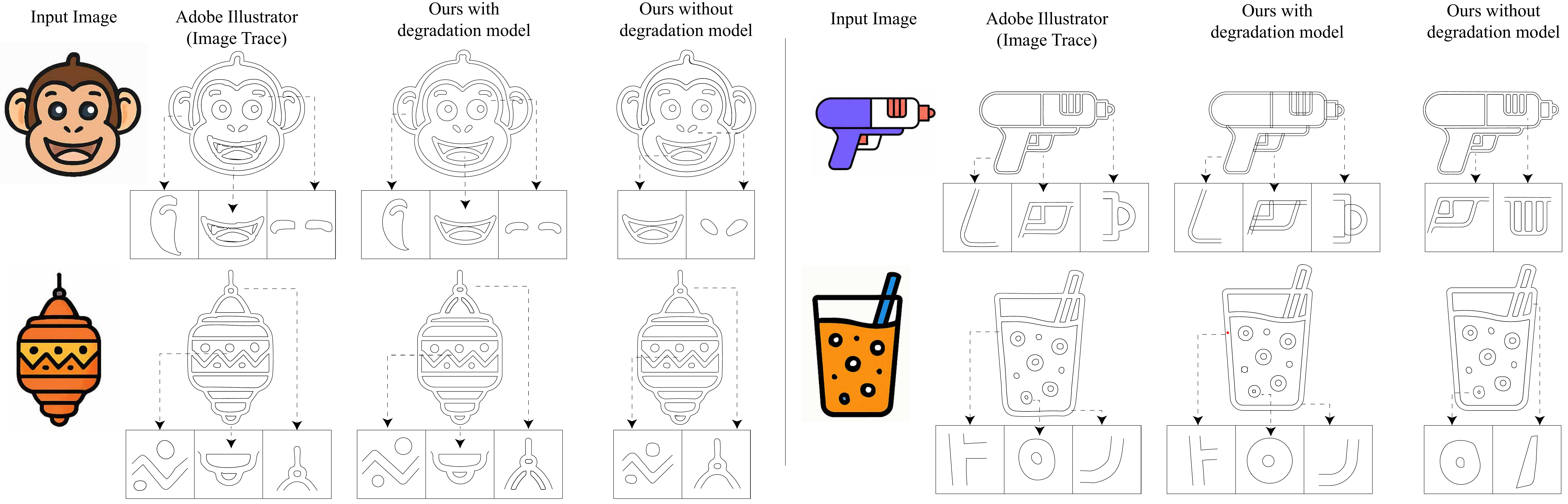}
\vspace{-3mm}
\caption{\textbf{Quality comparison with and without degradation model.} Four test cases (arranged in 2×2 grid) comparing vectorization quality across different methods. For each input image, we show generated outlines from: \textbf{Adobe Illustrator} (Image Trace), \textbf{Ours with degradation model}, and \textbf{Ours without degradation model}. Adobe Illustrator introduces artifacts and geometric irregularities (highlighted in boxes). Our method with noise-aware training substantially mitigates these artifacts, producing cleaner, well-regularized geometry suitable for editing. Without such training, the model tends to replicate input noise and artifacts, failing to achieve comparable geometric quality and editability. Notably, in the water gun example, our method reconstructs smooth curves and decomposes the complex shell into smaller, easily editable components.}
\label{fig:noise_ablation}
\end{figure*}

\paragraph{Ablation on Degradation Model}
To assess the effectiveness of our degradation model, we conduct an ablation study comparing our full model against an identical configuration trained without the degradation pipeline. 
All test inputs are generated by text-to-image models (GPT-4o~\cite{openai2024gpt4o} and Gemini~\cite{gemini2025gemini2_5}), representing practical deployment scenarios where raster graphics often contain artifacts and imperfections. 
As illustrated in Figure~\ref{fig:noise_ablation}, our degradation pipeline enables the model to produce significantly cleaner and more regularized geometry compared to both traditional vectorization (Adobe Illustrator's Image Trace) and the baseline variant trained without training-time degradation simulation. 
The model trained without degradation model tends to reproduce input artifacts, limiting geometric quality and editability, whereas our approach demonstrates robust artifact suppression and improved structural coherence. 

\paragraph{Ablation on Test-Time Scaling}

\begin{figure}[H]
\centering
\includegraphics[width=0.98\columnwidth]{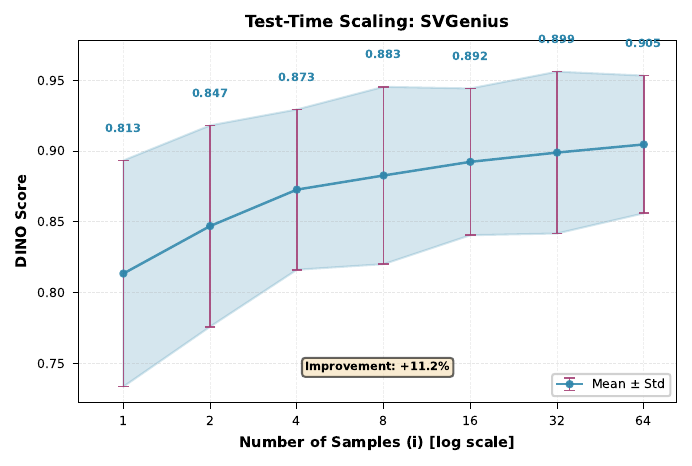}
\caption{\textbf{Test-time scaling analysis on SVGenius.} DinoScore as a function of parallel generation count, averaged over three random seeds. Increasing parallel samples improves reconstruction quality through diversity sampling.}
\label{fig:test_time_scaling}
\end{figure}

We investigate the effect of test-time scaling~\cite{snell2024scaling} by varying the number of parallel generations and selecting the best output based on DinoScore. For each configuration, we generate $n$ samples in parallel and select the highest-quality output.
As shown in Figure~\ref{fig:test_time_scaling}, the quality improves with the number of parallel generations, demonstrating the benefit of test-time scaling. 
While additional samples continue to improve quality, the gains become marginal beyond a certain threshold, indicating diminishing returns relative to computational cost. This trade-off allows users to balance quality requirements against inference budget in deployment scenarios.

\paragraph{Ablation on Outline Raster Representation.}
We compare model performance when conditioned on \emph{colored} versus \emph{outline-only} inputs. Using identical architecture and training configuration, we train another variant on RGB inputs. Quantitative evaluation is presented in Table~\ref{tab:ablation_outline}.
Outline conditioning consistently outperforms color-based training across all metrics, benchmarks, and difficulty levels, with particularly pronounced gains on harder samples. 
Pre-vectorizing the input to extract outlines acts as an effective normalization step, eliminating appearance-related training-testing mismatch and allowing the model to focus on geometric structure. 
This leads to improved shape coherence and topological consistency in the generated SVGs.

\begin{table}[t]
\centering
\caption{Ablation study comparing colored vs. outline-only input conditioning using the same model architecture and training setup. We report results across SArena and SVGenius benchmarks at three difficulty levels. Bold indicates better performance.}
\label{tab:ablation_outline}
\resizebox{\columnwidth}{!}{
\begin{tabular}{@{}l|cccc@{}}
\toprule
\textbf{Dataset/Difficulty} & \textbf{SSIM $\uparrow$} & \textbf{LPIPS $\downarrow$} & \textbf{MSE $\downarrow$} & \textbf{Dino $\uparrow$} \\
\midrule
\multicolumn{5}{l}{\textit{SArena Icon Generation}} \\
\midrule
Easy (Color) & 0.878 & 0.055 & 0.031 & 0.978 \\
Easy (Outline) & \textbf{0.937} & \textbf{0.031} & \textbf{0.011} & \textbf{0.992} \\
\midrule
Medium (Color) & 0.793 & 0.091 & 0.023 & 0.953 \\
Medium (Outline) & \textbf{0.895} & \textbf{0.058} & \textbf{0.013} & \textbf{0.981} \\
\midrule
Hard (Color) & 0.715 & 0.162 & 0.033 & 0.928 \\
Hard (Outline) & \textbf{0.857} & \textbf{0.093} & \textbf{0.022} & \textbf{0.975} \\
\midrule
\multicolumn{5}{l}{\textit{SVGenius}} \\
\midrule
Easy (Color) & 0.906 & 0.061 & 0.024 & 0.983 \\
Easy (Outline) & \textbf{0.944} & \textbf{0.028} & \textbf{0.008} & \textbf{0.995} \\
\midrule
Medium (Color) & 0.761 & 0.118 & 0.023 & 0.947 \\
Medium (Outline) & \textbf{0.868} & \textbf{0.08} & \textbf{0.015} & \textbf{0.977} \\
\midrule
Hard (Color) & 0.697 & 0.165 & 0.032 & 0.887 \\
Hard (Outline) & \textbf{0.83} & \textbf{0.12} & \textbf{0.023} & \textbf{0.958} \\
\bottomrule
\end{tabular}
}
\end{table}

%% file: sec/X_suppl.tex
\clearpage
\setcounter{page}{1}
\maketitlesupplementary

\section{Additional Results}
\label{sec:additional_results}
\begin{figure}[h]
\centering
\includegraphics[width=1\columnwidth]{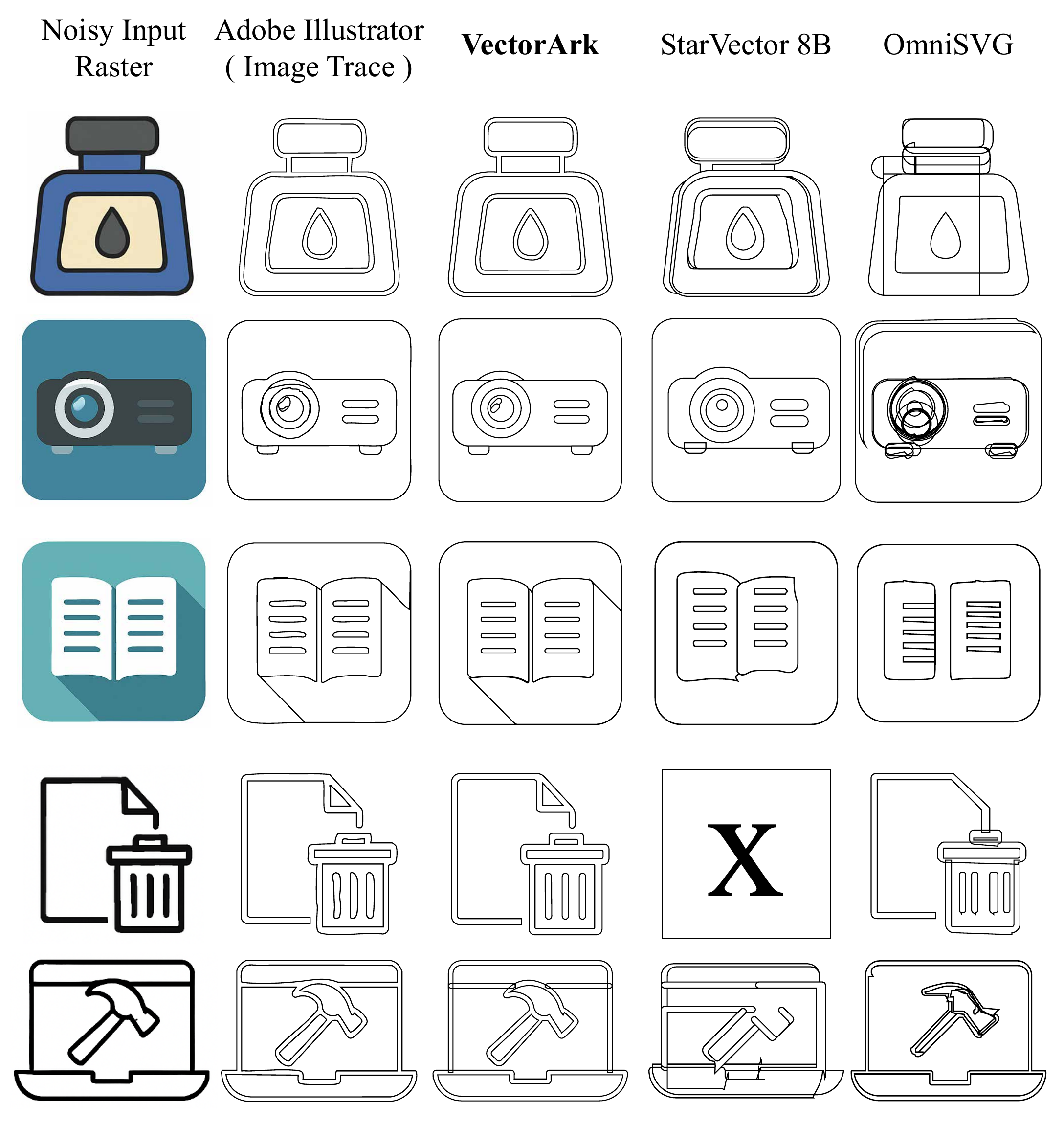}
\caption{Outline comparison across methods on noisy inputs. \textit{(Zoom in to see geometric artifacts.)}}
\label{fig:outline_comparison}
\end{figure}

\rebuttal{\subsection{Outline Comparison on Noisy Inputs}}
Figure~\ref{fig:outline_comparison} compares the generated outlines on noisy inputs. Our method maintains cleaner geometric structure compared to competing approaches.

\rebuttal{\subsection{Vectorizer Comparison}}
\rebuttal{Our preprocessing pipeline converts a raster image into an outline representation before feeding it to the model. While the main paper uses Adobe Illustrator's Image Trace for this step, we investigated several standard open-source alternatives to assess pipeline robustness and identify a freely available replacement.}

\begin{figure}[h]
\centering
\vspace{-2mm}
\begin{tabular}{@{}c@{\hfill}c@{\hfill}c@{\hfill}c@{}}
\includegraphics[width=0.22\linewidth]{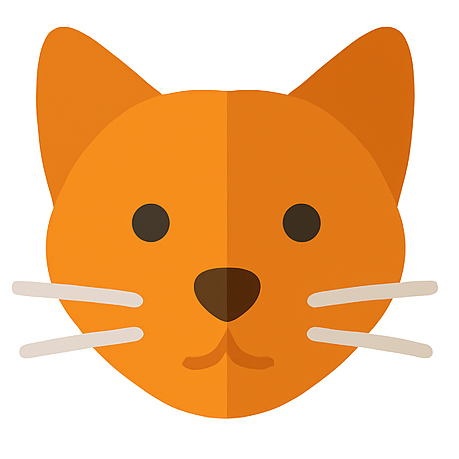} &
\includegraphics[width=0.22\linewidth]{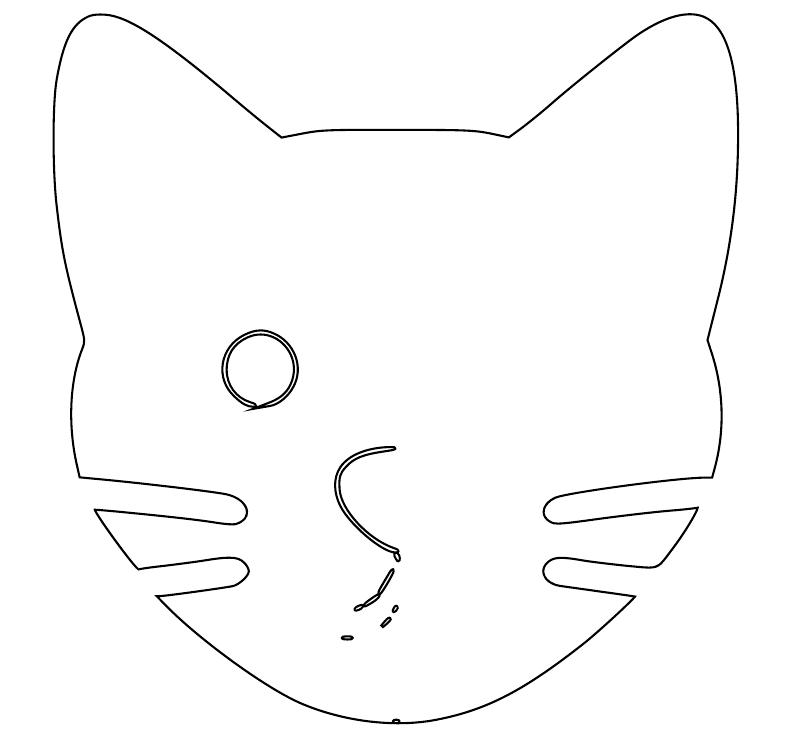} &
\includegraphics[width=0.22\linewidth]{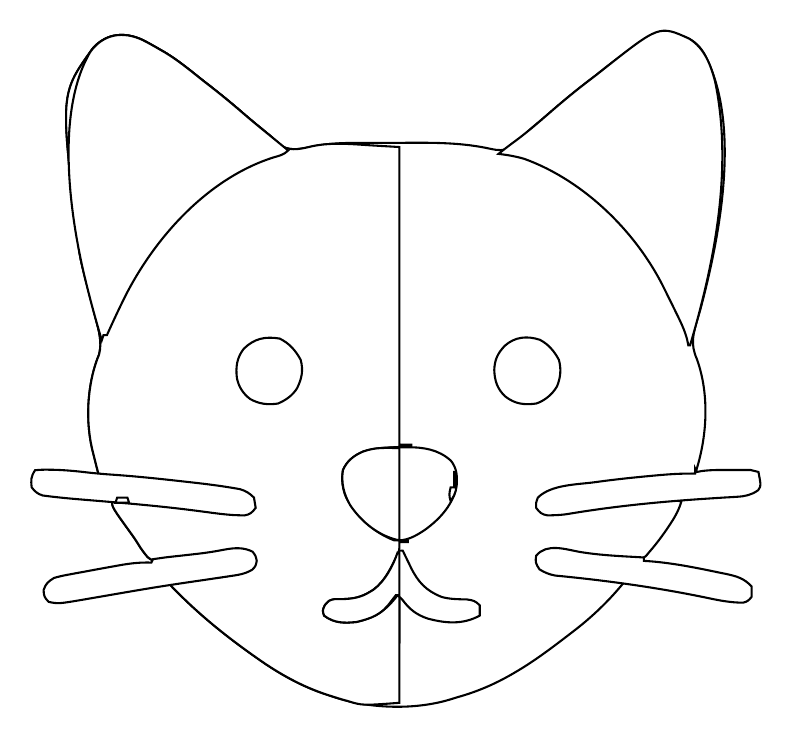} &
\includegraphics[width=0.22\linewidth]{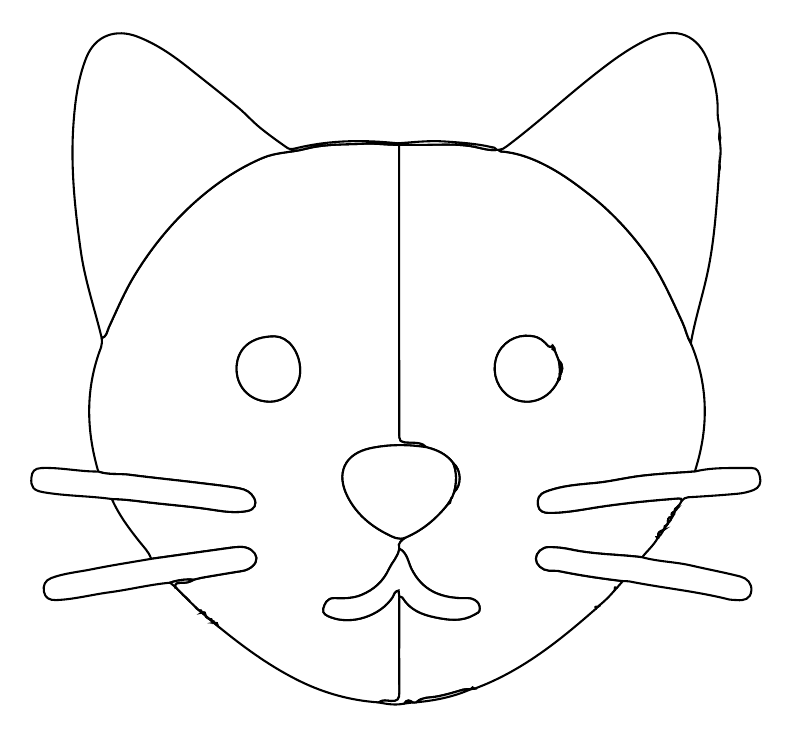} \\
(a) Input & (b) Potrace & (c) VTracer & (d) Image Trace \\[4pt]
\multicolumn{2}{c}{\includegraphics[width=0.22\linewidth]{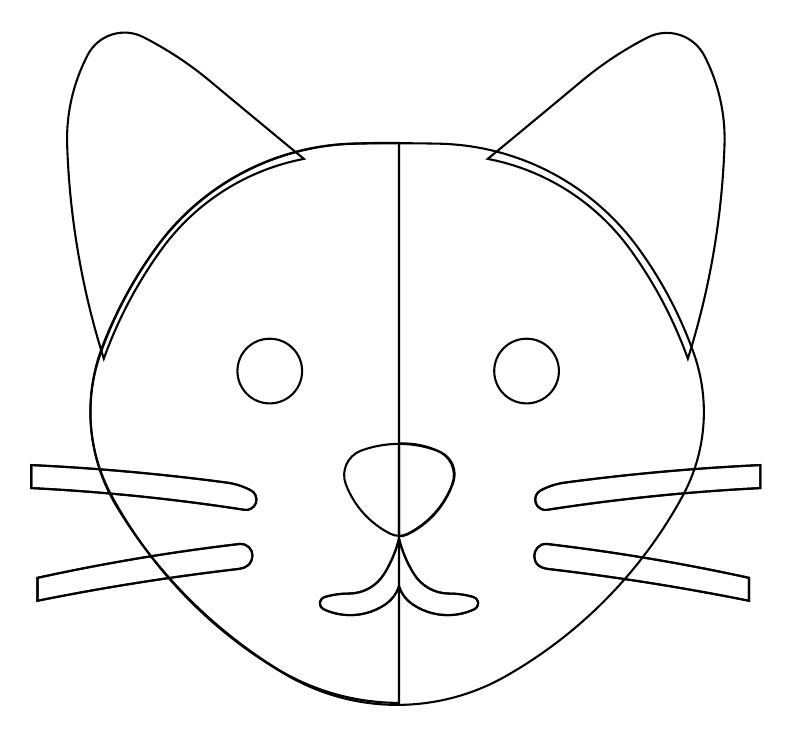}} &
\multicolumn{2}{c}{\includegraphics[width=0.22\linewidth]{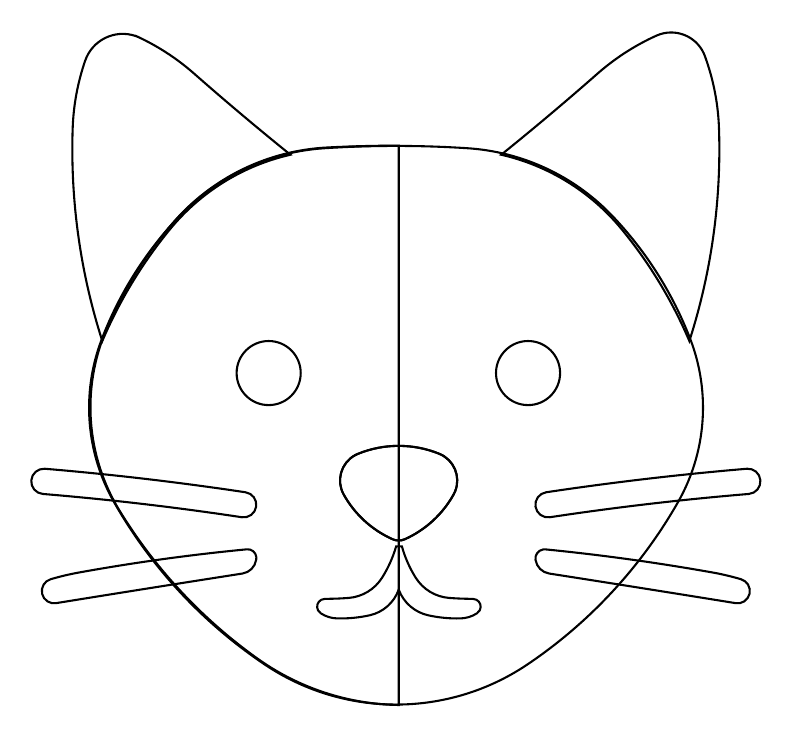}} \\
\multicolumn{2}{c}{(e) Ours (VTracer input)} &
\multicolumn{2}{c}{(f) Ours (Image Trace input)}
\end{tabular}
\vspace{-2mm}
\rebuttal{\caption{Vectorizer comparison and pipeline robustness on a sample input.}}
\label{fig:supp_vtracer}
\vspace{-3mm}
\end{figure}

\rebuttal{\paragraph{Evaluated Vectorizers.}
We experimented with three widely used vectorizers: Potrace~\cite{Potrace}, VTracer~\cite{vtracer2020}, and the optimization-based method LIVE~\cite{xu2022live}. As shown in the top row of Figure~\ref{fig:supp_vtracer}, Potrace produces patchy, often broken outlines with low noise resistance, making it unsuitable for practical use. LIVE requires manually specifying presets and the number of paths, adding significant user complexity; despite taking $>$10 minutes per image, it failed to produce meaningful results on most of the easy-category examples. Among the alternatives, VTracer yielded the most consistent and usable outlines. Although its outputs contain visible artifacts compared to Image Trace (Figure~\ref{fig:supp_vtracer}, top row), the overall structure is well preserved.}

\rebuttal{\paragraph{VTracer as a Drop-In Replacement.}
We additionally evaluated using VTracer in place of Image Trace in the full preprocessing pipeline. As shown in the bottom row of Figure~\ref{fig:supp_vtracer}, despite the noisier outline input, the final SVG quality remains comparable. Quantitatively, both vectorizers achieve a DINO score of approximately $0.98$ on SVGenius, demonstrating that our pipeline is robust to the choice of vectorizer and that VTracer is a viable open-source alternative.}

\begin{figure}[h]
\centering
\vspace{-2mm}
\includegraphics[width=0.8\columnwidth]{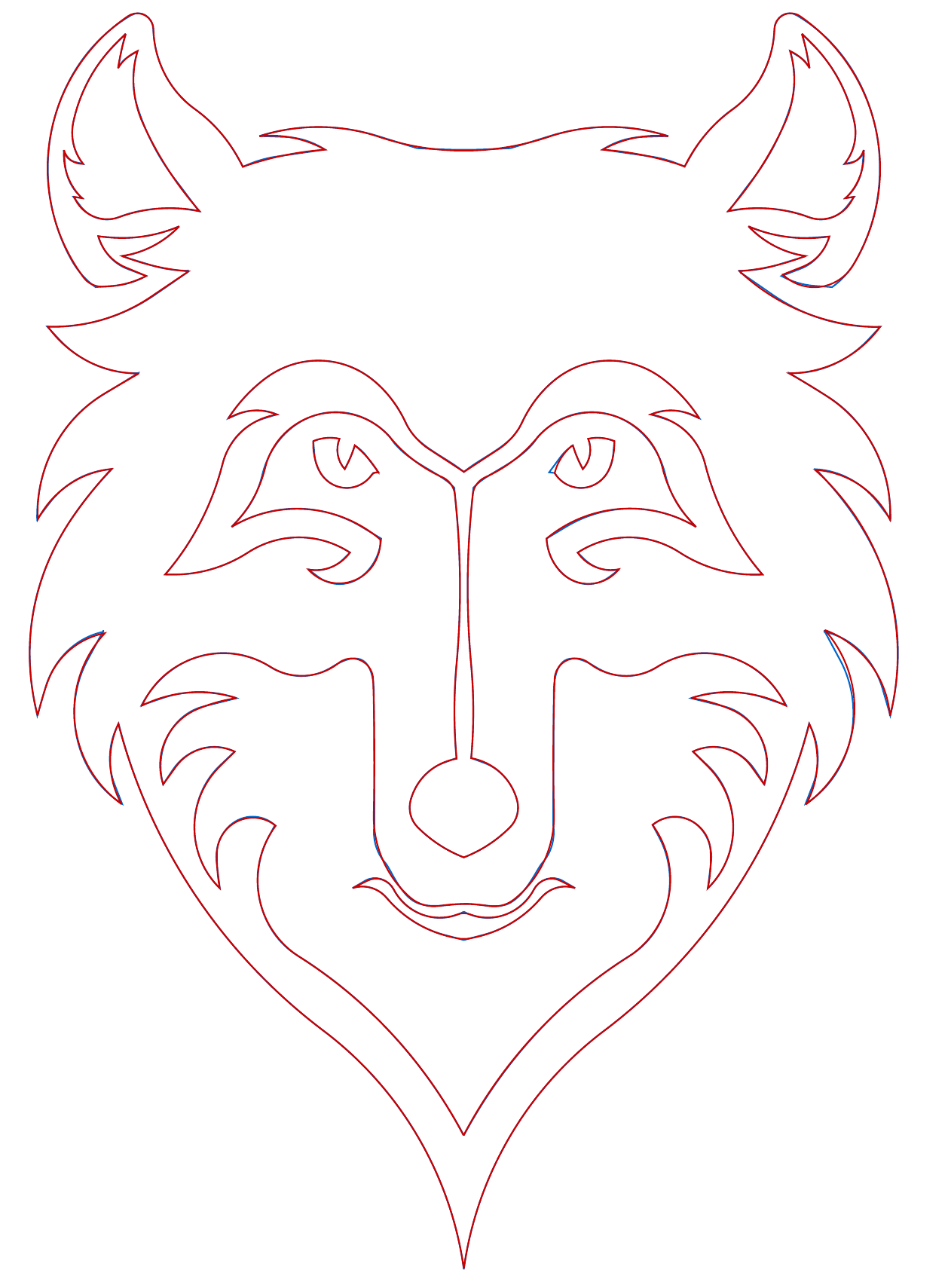}
\vspace{-2mm}
\caption{Intricate reconstruction example. Blue: original SVG; red: our reconstruction.}
\label{fig:supp_intricate}
\vspace{-3mm}
\end{figure}

\rebuttal{\subsection{Intricate Reconstruction}}
\rebuttal{Figure~\ref{fig:supp_intricate} highlights the expressive power of the rounded polygon representation on a highly intricate vector graphic: a public-domain wolf-head silhouette\footnote{\url{https://freesvg.org/wolf-head-silhouette}} with roughly $300$ control points and dense local structure. Even at this level of geometric complexity, the reconstruction remains closely aligned with the original SVG while requiring substantially fewer tokens than raw SVG commands. This example shows that the proposed representation is not restricted to simple icon-like geometry, but can also support detailed and visually complex designs in a compact form.}

\begin{figure}[h]
\centering
\vspace{-2mm}
\begin{tabular}{@{}c@{\hfill}c@{}}
\includegraphics[width=0.42\linewidth]{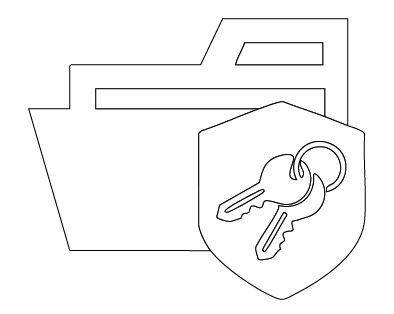} &
\includegraphics[width=0.42\linewidth]{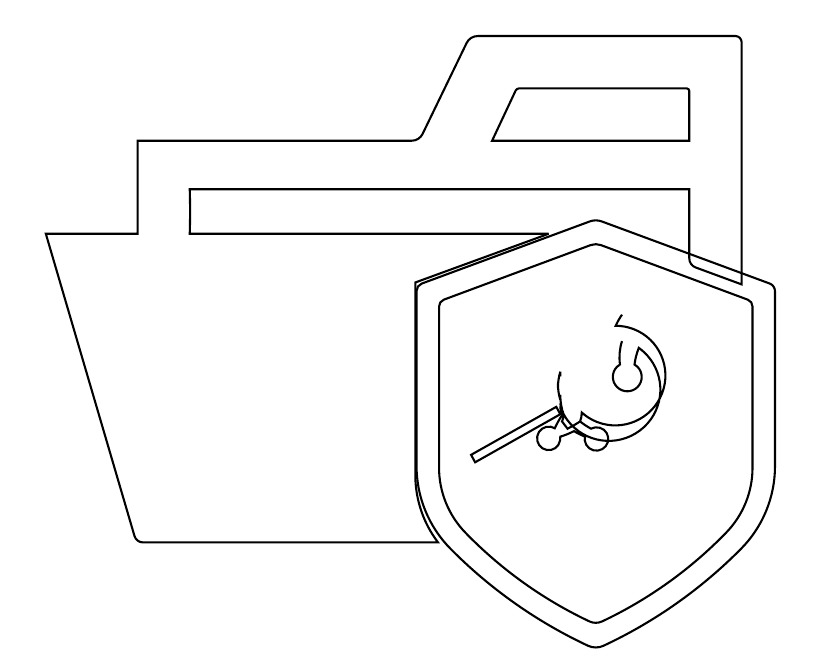} \\
(a) Vectorized input & (b) Model output
\end{tabular}
\vspace{-2mm}
\rebuttal{\caption{Failure case with dense local path structure concentrated in a small region.}}
\label{fig:supp_failure}
\vspace{-3mm}
\end{figure}

\rebuttal{\subsection{Failure Case}}
\rebuttal{Figure~\ref{fig:supp_failure} shows a representative failure mode. When many paths are concentrated in a very small spatial region, the model struggles to resolve fine-grained local topology and tends to merge or misplace individual primitives. We observe that increasing the input raster resolution alleviates this issue and produces noticeably better reconstructions in such cases.}

\rebuttal{\subsection{Additional Metrics and Analysis}}
\label{sec:additional_analysis}

\rebuttal{To further substantiate our claims, we report complementary evaluations beyond those in the main paper: results without test-time scaling, a geometry-level metric (Chamfer distance), a qualitative analysis of reconstruction behavior on easy-category samples, and an inference-time comparison.}

\rebuttal{\subsubsection{Evaluation Without Test-Time Scaling}}
\rebuttal{Table~\ref{tab:no_tts} reports performance with a single stochastic sample ($N{=}1$), \ie without best-of-$N$ selection. Our method achieves the best scores across all four metrics, maintaining a clear lead even without test-time scaling.}

\begin{table}[h]
    \centering
    \vspace{-2mm}
    \caption{Performance without test-time scaling (N=1) on SVGenius (averaged across Easy/Medium/Hard).}
    \label{tab:no_tts}
    \resizebox{0.95\columnwidth}{!}{
    \begin{tabular}{@{}l|ccccc@{}}
    \toprule
    \textbf{Metric} & \textbf{GPT-4o} & \textbf{Gemini} & \textbf{OmniSVG} & \textbf{StarVector} & \textbf{Ours} \\
    \midrule
    SSIM $\uparrow$ & 0.590 & 0.540 & 0.679 & 0.724 & \textbf{0.815} \\
    LPIPS $\downarrow$ & 0.274 & 0.330 & 0.232 & 0.226 & \textbf{0.136} \\
    MSE $\downarrow$ & 0.094 & 0.124 & 0.066 & 0.065 & \textbf{0.033} \\
    DINO $\uparrow$ & 0.929 & 0.886 & 0.885 & 0.862 & \textbf{0.940} \\
    \bottomrule
    \end{tabular}
    }
    \vspace{-2mm}
\end{table}


\rebuttal{\subsubsection{Chamfer Distance}}
\rebuttal{Beyond raster-space evaluation, we additionally report a metric that operates directly in the vector domain and is better suited to assessing geometric reconstruction quality. Table~\ref{tab:supp_chamfer} reports Chamfer distance between predicted and ground-truth control points, a metric better aligned with the geometric nature of SVG outputs. Our method achieves the lowest Chamfer distance on Medium and Hard splits, and matches OmniSVG on Easy, reinforcing the strong reconstruction quality observed with raster-space metrics.}

\begin{table}[h]
\centering
\caption{\rebuttal{Chamfer distance ($\downarrow$) between predicted and ground-truth control points.}}
\label{tab:supp_chamfer}
\resizebox{\columnwidth}{!}{
\begin{tabular}{@{}l|ccccc@{}}
\toprule
\textbf{Difficulty} & \textbf{GPT-4o} & \textbf{Gemini} & \textbf{OmniSVG} & \textbf{StarVector} & \textbf{Ours} \\
\midrule
Easy & 0.067 & 0.080 & \textbf{0.024} & 0.055 & \textbf{0.024} \\
Medium & 0.080 & 0.093 & 0.063 & 0.108 & \textbf{0.054} \\
Hard & 0.081 & 0.086 & 0.073 & 0.133 & \textbf{0.071} \\
\bottomrule
\end{tabular}
}
\end{table}

\begin{figure}[h]
\centering
\vspace{-2mm}
\includegraphics[width=0.7\columnwidth]{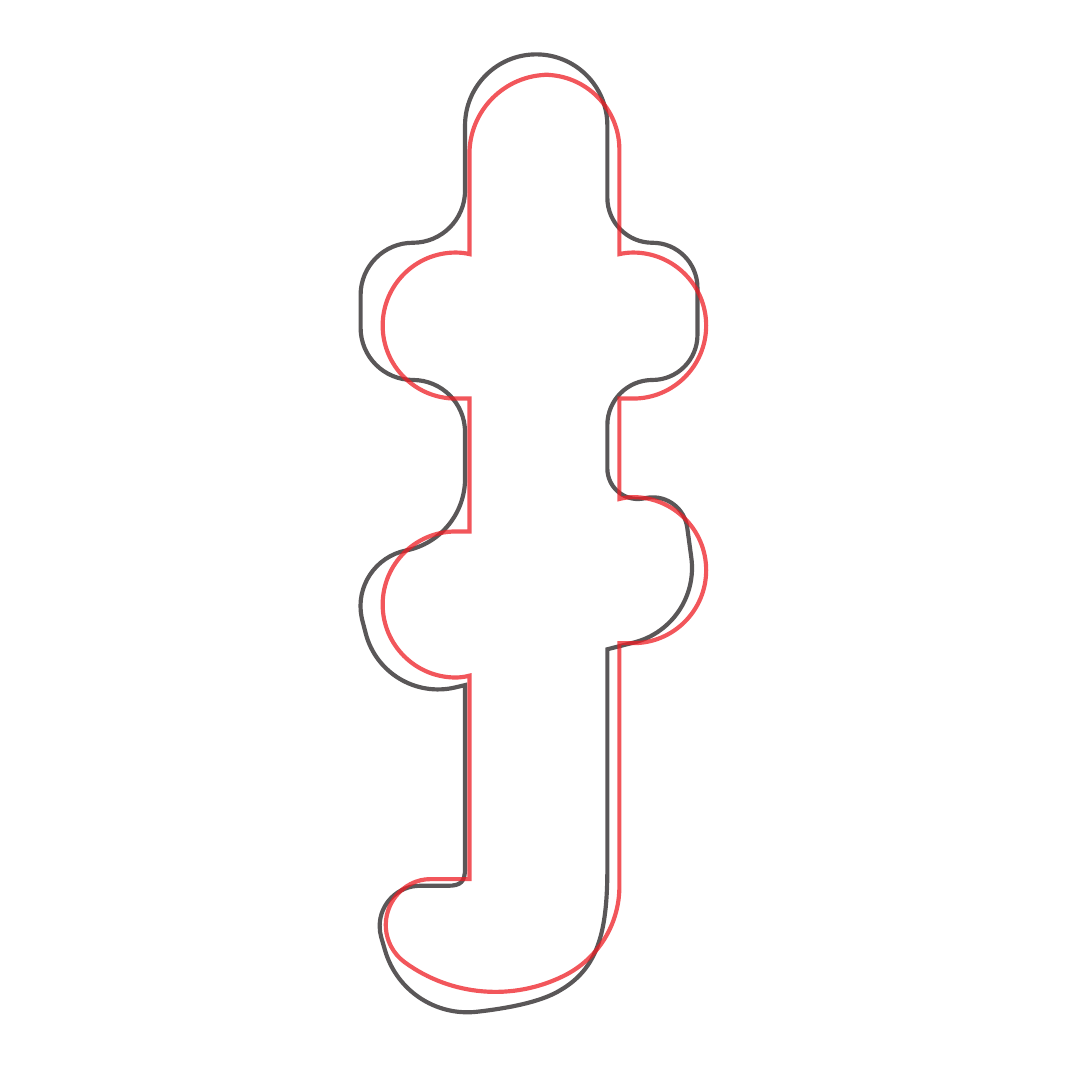}
\vspace{-2mm}
\rebuttal{\caption{Easy-category reconstruction: our output (red) versus ground truth (black). The rounded polygon representation smooths minor irregularities while preserving overall shape fidelity.}}
\label{fig:supp_easy_deviation}
\vspace{-3mm}
\end{figure}

\rebuttal{\subsubsection{Reconstruction Behavior on Easy Samples}}
\rebuttal{As noted in the ablation study, in the main paper, the raw SVG-command representation attains slightly stronger pixel-level metrics on the Easy split while our rounded polygon representation leads in semantic similarity (DINO). Figure~\ref{fig:supp_easy_deviation} illustrates why: our representation regularizes minor local irregularities, so the predicted paths (red) may exhibit small geometric deviations from the ground truth (black). On easy samples, which contain only a few primitives, each such deviation has a proportionally larger effect on pixel-level metrics. However, DINO---which captures higher-level shape semantics---remains equal or better, indicating that the overall structure is faithfully preserved. On Medium and Hard samples this effect is less pronounced, as the dominant source of error shifts to broader geometric misalignment where our representation's stability becomes the decisive advantage.}

\rebuttal{\subsubsection{Inference Time}}
\rebuttal{On an A100, our method generates one SVG in $3$--$4$\,s on easy-category samples, compared with $8$--$16$\,s for StarVector and $4$--$6$\,s for OmniSVG. The compact 1B architecture (vs.\ 8B for StarVector and 3B for OmniSVG) enables both faster generation and lower memory consumption. Classical image vectorization adds less than $1$\,s at $1024{\times}1024$ resolution, and batched DINO scoring for test-time scaling adds $<0.25$\,s. The same vectorized outline can be reused across multiple stochastic samples, further amortizing preprocessing cost.}

\rebuttal{\subsection{Extended Benchmark Evaluation}}
\rebuttaldel{\paragraph{Setup.}}
\rebuttal{\subsubsection{Setup}}
We evaluate on ten benchmarks from three recent works: eight from StarVector~\cite{Rodriguez_StarVector_2025} (\textbf{SVG-Stack}, \textbf{SVG-Stack-Simple}, \textbf{SVG-Fonts}, \textbf{SVG-Fonts-Simple}, \textbf{SVG-Icons}, \textbf{SVG-Icons-Simple}, \textbf{SVG-Emoji}, \textbf{SVG-Emoji-Simple}), as well as \textbf{MMSVG-Icon}~\cite{yang2025omnisvg} and \textbf{SVGX\_core\_250k}~\cite{xing2024llm4svg}. These benchmarks encompass diverse vector graphic domains including fonts, icons, emoji, and multi-colored graphics. To systematically assess performance across geometric complexity, we categorize samples into three difficulty tiers based on path primitive counts: \emph{Easy} ($< 64$), \emph{Medium} ($64$--$128$), and \emph{Hard} ($> 128$).

\rebuttaldel{\paragraph{Quantitative Results.}}
\rebuttal{\subsubsection{Quantitative Results}}
Tables \ref{tab:nonsimplified_results}, \ref{tab:simplified_results}, and \ref{tab:other_benchmarks} present comprehensive quantitative comparisons across all benchmarks. Our method consistently achieves state-of-the-art performance across the majority of datasets, metrics, and complexity tiers.

On non-simplified datasets (\emph{SVG-Stack}, \emph{SVG-Fonts}, \emph{SVG-Emoji}, \emph{MMSVG-Icon}, \emph{SVGX\_core\_250k}), our method ranks first across all four evaluation metrics on the majority of complexity tiers. The performance advantages are particularly pronounced on higher complexity levels, where our geometry-first approach demonstrates stronger capability in reconstructing intricate path topology.

On simplified datasets, the results remain highly competitive. For \emph{SVG-Emoji-Simple} and \emph{SVG-Fonts-Simple}, we lead across all metrics and complexity levels. On \emph{SVG-Icons} and \emph{SVG-Icons-Simple}, while OmniSVG achieves competitive perceptual similarity (LPIPS) on certain subsets, our method maintains stronger structural fidelity (SSIM) and competitive semantic coherence (DINO).

\begin{table*}[h]
\centering
\caption{Quantitative comparison on non-simplified StarVector benchmarks. Best results in \textbf{bold}, second best \underline{underlined}.}
\label{tab:nonsimplified_results}
\footnotesize
\setlength{\tabcolsep}{3.5pt}
\begin{tabular}{ll|ccc|ccc|ccc|ccc}
\toprule
\multirow{2}{*}{\textbf{Method}} & \multirow{2}{*}{\textbf{Metric}} & \multicolumn{3}{c|}{\textbf{SVG-Stack}} & \multicolumn{3}{c|}{\textbf{SVG-Fonts}} & \multicolumn{3}{c|}{\textbf{SVG-Icons}} & \multicolumn{3}{c}{\textbf{SVG-Emoji}} \\
& & Easy & Med & Hard & Easy & Med & Hard & Easy & Med & Hard & Easy & Med & Hard \\
\midrule
\multirow{4}{*}{OmniSVG} 
& LPIPS$\downarrow$ & 0.2297 & 0.2572 & 0.3156 & \underline{0.0515} & \underline{0.0920} & \underline{0.1524} & \underline{0.0670} & \underline{0.1287} & \textbf{0.1734} & 0.2706 & 0.4266 & 0.4275 \\
& SSIM$\uparrow$ & 0.6128 & 0.6230 & 0.5787 & \underline{0.8962} & \underline{0.8238} & \underline{0.7323} & 0.8507 & 0.7658 & 0.7103 & 0.6633 & 0.5304 & 0.4930 \\
& DINO$\uparrow$ & 0.8976 & 0.8835 & 0.8523 & 0.9743 & \underline{0.9468} & \underline{0.9183} & \underline{0.9755} & \underline{0.9473} & \textbf{0.9339} & 0.9023 & 0.8621 & 0.8461 \\
& MSE$\downarrow$ & 0.1225 & 0.0839 & 0.1081 & \underline{0.0269} & \underline{0.0571} & \underline{0.0970} & \underline{0.0346} & \underline{0.0642} & \underline{0.0879} & 0.0749 & 0.1100 & 0.1201 \\
\midrule
\multirow{4}{*}{GPT-4o} 
& LPIPS$\downarrow$ & 0.2287 & 0.2452 & 0.2728 & 0.1579 & 0.1944 & 0.2393 & 0.1637 & 0.2069 & 0.2330 & 0.2423 & 0.3488 & 0.3891 \\
& SSIM$\uparrow$ & 0.5806 & 0.6120 & 0.5825 & 0.7513 & 0.7089 & 0.6287 & 0.7543 & 0.6898 & 0.6560 & 0.6678 & 0.5672 & 0.5253 \\
& DINO$\uparrow$ & 0.9541 & 0.9399 & 0.9238 & 0.9647 & 0.9326 & 0.9001 & 0.9565 & 0.9161 & 0.9111 & 0.9508 & 0.8975 & 0.8336 \\
& MSE$\downarrow$ & 0.1447 & 0.0942 & 0.1168 & 0.1186 & 0.1258 & 0.1650 & 0.0682 & 0.0973 & 0.1191 & 0.0719 & 0.1060 & 0.1070 \\
\midrule
\multirow{4}{*}{Gemini-2.5-Pro} 
& LPIPS$\downarrow$ & 0.2537 & 0.2913 & 0.2795 & 0.1786 & 0.2077 & 0.2431 & 0.2070 & 0.2417 & 0.2586 & 0.2946 & 0.4001 & 0.3942 \\
& SSIM$\uparrow$ & 0.5370 & 0.5673 & 0.5704 & 0.7280 & 0.6950 & 0.6298 & 0.6971 & 0.6309 & 0.6136 & 0.6201 & 0.5280 & 0.4870 \\
& DINO$\uparrow$ & 0.9485 & 0.9328 & \underline{0.9310} & 0.9384 & 0.9036 & 0.9086 & 0.9374 & 0.9249 & 0.9138 & 0.9385 & 0.9066 & \underline{0.9013} \\
& MSE$\downarrow$ & 0.1740 & 0.1183 & 0.1109 & 0.1322 & 0.1448 & 0.1648 & 0.1063 & 0.1394 & 0.1436 & 0.0917 & 0.1174 & 0.1087 \\
\midrule
\multirow{4}{*}{StarVector\_8B} 
& LPIPS$\downarrow$ & \underline{0.0609} & \underline{0.1166} & \underline{0.1933} & 0.0624 & 0.1194 & 0.1961 & 0.0764 & 0.1383 & 0.2146 & \underline{0.0857} & \underline{0.2683} & \underline{0.3605} \\
& SSIM$\uparrow$ & \underline{0.8714} & \underline{0.8114} & \underline{0.7046} & 0.8754 & 0.7978 & 0.7069 & \underline{0.8532} & \underline{0.7712} & \underline{0.7109} & \underline{0.8503} & \underline{0.6588} & \underline{0.5883} \\
& DINO$\uparrow$ & \underline{0.9774} & \underline{0.9503} & 0.9180 & \underline{0.9756} & 0.9192 & 0.8489 & 0.9606 & 0.9261 & 0.8476 & \underline{0.9875} & \underline{0.9200} & 0.8411 \\
& MSE$\downarrow$ & \underline{0.0275} & \underline{0.0357} & \underline{0.0766} & 0.0396 & 0.0694 & 0.1285 & 0.0357 & 0.0669 & 0.0961 & \underline{0.0220} & \underline{0.0804} & \underline{0.0886} \\
\midrule
\multirow{4}{*}{\textbf{Ours}} 
& LPIPS$\downarrow$ & \textbf{0.0274} & \textbf{0.0577} & \textbf{0.0643} & \textbf{0.0217} & \textbf{0.0277} & \textbf{0.0555} & \textbf{0.0565} & \textbf{0.0831} & \underline{0.1845} & \textbf{0.0282} & \textbf{0.0709} & \textbf{0.1517} \\
& SSIM$\uparrow$ & \textbf{0.9531} & \textbf{0.9150} & \textbf{0.9146} & \textbf{0.9651} & \textbf{0.9505} & \textbf{0.9074} & \textbf{0.9249} & \textbf{0.8683} & \textbf{0.7193} & \textbf{0.9588} & \textbf{0.8946} & \textbf{0.7971} \\
& DINO$\uparrow$ & \textbf{0.9905} & \textbf{0.9744} & \textbf{0.9791} & \textbf{0.9912} & \textbf{0.9868} & \textbf{0.9773} & \textbf{0.9814} & \textbf{0.9746} & \underline{0.9259} & \textbf{0.9917} & \textbf{0.9841} & \textbf{0.9602} \\
& MSE$\downarrow$ & \textbf{0.0044} & \textbf{0.0064} & \textbf{0.0111} & \textbf{0.0075} & \textbf{0.0073} & \textbf{0.0209} & \textbf{0.0166} & \textbf{0.0323} & \textbf{0.0769} & \textbf{0.0043} & \textbf{0.0142} & \textbf{0.0288} \\
\bottomrule
\end{tabular}
\end{table*}

\begin{table*}[h]
\centering
\caption{Quantitative comparison on simplified StarVector benchmarks. Best results in \textbf{bold}, second best \underline{underlined}.}
\label{tab:simplified_results}
\footnotesize
\setlength{\tabcolsep}{3.5pt}
\begin{tabular}{ll|ccc|ccc|ccc|ccc}
\toprule
\multirow{2}{*}{\textbf{Method}} & \multirow{2}{*}{\textbf{Metric}} & \multicolumn{3}{c|}{\textbf{SVG-Stack-Simple}} & \multicolumn{3}{c|}{\textbf{SVG-Fonts-Simple}} & \multicolumn{3}{c|}{\textbf{SVG-Icons-Simple}} & \multicolumn{3}{c}{\textbf{SVG-Emoji-Simple}} \\
& & Easy & Med & Hard & Easy & Med & Hard & Easy & Med & Hard & Easy & Med & Hard \\
\midrule
\multirow{4}{*}{OmniSVG} 
& LPIPS$\downarrow$ & 0.0911 & 0.1712 & \underline{0.2201} & 0.0766 & 0.1748 & 0.2240 & \underline{0.0736} & \textbf{0.1202} & \textbf{0.1453} & \underline{0.0922} & 0.2920 & 0.3416 \\
& SSIM$\uparrow$ & 0.7203 & 0.6078 & 0.5698 & 0.8196 & 0.6992 & 0.6530 & 0.8462 & 0.7976 & \underline{0.7603} & \underline{0.7451} & 0.4376 & 0.4226 \\
& DINO$\uparrow$ & 0.9696 & \underline{0.9308} & 0.8901 & 0.9659 & 0.9208 & 0.8787 & \underline{0.9714} & \underline{0.9493} & \underline{0.9362} & \underline{0.9709} & 0.8761 & 0.8257 \\
& MSE$\downarrow$ & 0.0919 & 0.1302 & 0.1706 & 0.0437 & 0.0830 & 0.1143 & 0.0370 & \underline{0.0578} & \underline{0.0781} & 0.0567 & 0.1954 & 0.1867 \\
\midrule
\multirow{4}{*}{GPT-4o} 
& LPIPS$\downarrow$ & 0.2049 & 0.2569 & 0.2552 & 0.2004 & 0.2465 & 0.2659 & 0.1553 & 0.1756 & 0.2009 & 0.2350 & 0.3491 & 0.3629 \\
& SSIM$\uparrow$ & 0.5978 & 0.5369 & 0.5780 & 0.7125 & 0.6365 & 0.6041 & 0.7660 & 0.7292 & 0.7025 & 0.5871 & 0.3867 & 0.4169 \\
& DINO$\uparrow$ & 0.9491 & 0.9294 & 0.8914 & 0.9358 & 0.9022 & 0.8921 & 0.9500 & 0.9451 & 0.9198 & 0.9122 & 0.8437 & 0.8030 \\
& MSE$\downarrow$ & 0.1100 & 0.1429 & 0.1334 & 0.0802 & 0.1229 & 0.1388 & 0.0631 & 0.0872 & 0.1036 & 0.1367 & 0.1893 & 0.1905 \\
\midrule
\multirow{4}{*}{Gemini-2.5-Pro} 
& LPIPS$\downarrow$ & 0.2535 & 0.2781 & 0.2784 & 0.2004 & 0.2569 & 0.2682 & 0.2049 & 0.2362 & 0.2305 & 0.2754 & 0.3720 & 0.3861 \\
& SSIM$\uparrow$ & 0.5497 & 0.5072 & 0.5319 & 0.6963 & 0.6123 & 0.5987 & 0.7049 & 0.6507 & 0.6578 & 0.5322 & 0.3587 & 0.3599 \\
& DINO$\uparrow$ & 0.9259 & 0.9133 & \underline{0.9066} & 0.9195 & 0.8961 & \underline{0.9019} & 0.9308 & 0.9264 & 0.9258 & 0.8961 & \underline{0.8800} & \underline{0.8611} \\
& MSE$\downarrow$ & 0.1704 & 0.1822 & 0.1683 & 0.1217 & 0.1583 & 0.1501 & 0.1048 & 0.1351 & 0.1301 & 0.1865 & 0.2014 & 0.2338 \\
\midrule
\multirow{4}{*}{StarVector\_8B} 
& LPIPS$\downarrow$ & \underline{0.0824} & \underline{0.1679} & 0.2295 & \underline{0.0610} & \underline{0.1589} & \underline{0.2164} & \textbf{0.0686} & 0.1338 & 0.1727 & 0.1094 & \underline{0.2865} & \underline{0.3409} \\
& SSIM$\uparrow$ & \underline{0.7626} & \underline{0.6384} & \underline{0.6294} & \underline{0.8425} & \underline{0.7058} & \underline{0.6547} & \underline{0.8582} & \underline{0.7980} & 0.7598 & 0.7436 & \underline{0.4953} & \underline{0.4576} \\
& DINO$\uparrow$ & \textbf{0.9849} & 0.9162 & 0.8465 & \underline{0.9815} & \underline{0.9234} & 0.8828 & 0.9690 & 0.9116 & 0.8767 & 0.9620 & 0.8250 & 0.7803 \\
& MSE$\downarrow$ & \underline{0.0541} & \underline{0.1038} & \underline{0.1052} & \underline{0.0349} & \underline{0.0747} & \underline{0.1075} & \underline{0.0331} & 0.0631 & 0.0823 & \underline{0.0561} & \underline{0.1132} & \underline{0.1441} \\
\midrule
\multirow{4}{*}{\textbf{Ours}} 
& LPIPS$\downarrow$ & \textbf{0.0390} & \textbf{0.0425} & \textbf{0.0574} & \textbf{0.0339} & \textbf{0.0787} & \textbf{0.1674} & 0.0798 & \underline{0.1209} & \underline{0.1697} & \textbf{0.0623} & \textbf{0.1008} & \textbf{0.2740} \\
& SSIM$\uparrow$ & \textbf{0.9460} & \textbf{0.9361} & \textbf{0.9246} & \textbf{0.9435} & \textbf{0.8758} & \textbf{0.7492} & \textbf{0.8995} & \textbf{0.8188} & \textbf{0.7606} & \textbf{0.9151} & \textbf{0.8465} & \textbf{0.6146} \\
& DINO$\uparrow$ & \underline{0.9823} & \textbf{0.9832} & \textbf{0.9653} & \textbf{0.9903} & \textbf{0.9729} & \textbf{0.9459} & \textbf{0.9722} & \textbf{0.9559} & \textbf{0.9400} & \textbf{0.9772} & \textbf{0.9593} & \textbf{0.8718} \\
& MSE$\downarrow$ & \textbf{0.0086} & \textbf{0.0103} & \textbf{0.0155} & \textbf{0.0111} & \textbf{0.0271} & \textbf{0.0545} & \textbf{0.0210} & \textbf{0.0478} & \textbf{0.0631} & \textbf{0.0129} & \textbf{0.0253} & \textbf{0.0758} \\
\bottomrule
\end{tabular}
\end{table*}

\begin{table*}[h]
\centering
\caption{Quantitative comparison on additional benchmarks. Best results in \textbf{bold}, second best \underline{underlined}.}
\label{tab:other_benchmarks}
\footnotesize
\setlength{\tabcolsep}{3.5pt}
\begin{tabular}{ll|ccc|ccc}
\toprule
\multirow{2}{*}{\textbf{Method}} & \multirow{2}{*}{\textbf{Metric}} & \multicolumn{3}{c|}{\textbf{MMSVG-Icon}} & \multicolumn{3}{c}{\textbf{SVGX\_core\_250k}} \\
& & Easy & Med & Hard & Easy & Med & Hard \\
\midrule
\multirow{4}{*}{OmniSVG} 
& LPIPS$\downarrow$ & 0.0758 & 0.1210 & \underline{0.1357} & 0.0651 & 0.1576 & 0.2513 \\
& SSIM$\uparrow$ & 0.8230 & 0.7581 & \underline{0.7337} & 0.8446 & 0.7091 & 0.5978 \\
& DINO$\uparrow$ & 0.9869 & 0.9711 & \underline{0.9665} & 0.9871 & 0.9552 & \underline{0.9242} \\
& MSE$\downarrow$ & 0.0277 & 0.0439 & \underline{0.0588} & 0.0300 & 0.0690 & 0.0883 \\
\midrule
\multirow{4}{*}{GPT-4o} 
& LPIPS$\downarrow$ & 0.2156 & 0.2602 & 0.2831 & 0.2055 & 0.2649 & 0.3237 \\
& SSIM$\uparrow$ & 0.6081 & 0.5579 & 0.5276 & 0.6299 & 0.5416 & 0.5193 \\
& DINO$\uparrow$ & 0.9737 & 0.9594 & 0.9403 & 0.9680 & 0.9434 & 0.8726 \\
& MSE$\downarrow$ & 0.1209 & 0.1271 & 0.1496 & 0.1408 & 0.1473 & 0.1259 \\
\midrule
\multirow{4}{*}{Gemini-2.5-Pro} 
& LPIPS$\downarrow$ & 0.2550 & 0.3220 & 0.3388 & 0.2322 & 0.3141 & 0.3438 \\
& SSIM$\uparrow$ & 0.5712 & 0.4945 & 0.4813 & 0.6004 & 0.5043 & 0.4711 \\
& DINO$\uparrow$ & 0.9466 & 0.9220 & 0.9171 & 0.9441 & 0.9261 & 0.9074 \\
& MSE$\downarrow$ & 0.1502 & 0.1649 & 0.1723 & 0.1647 & 0.1757 & 0.1540 \\
\midrule
\multirow{4}{*}{StarVector\_8B} 
& LPIPS$\downarrow$ & \underline{0.0451} & \underline{0.0964} & 0.1431 & \underline{0.0409} & \underline{0.1053} & \underline{0.2360} \\
& SSIM$\uparrow$ & \underline{0.9022} & \underline{0.8091} & 0.7292 & \underline{0.9101} & \underline{0.8086} & \underline{0.6559} \\
& DINO$\uparrow$ & \underline{0.9924} & \underline{0.9783} & 0.9551 & \underline{0.9910} & \underline{0.9625} & 0.9058 \\
& MSE$\downarrow$ & \underline{0.0151} & \underline{0.0433} & 0.0627 & \underline{0.0212} & \underline{0.0482} & \underline{0.0837} \\
\midrule
\multirow{4}{*}{\textbf{Ours}} 
& LPIPS$\downarrow$ & \textbf{0.0122} & \textbf{0.0179} & \textbf{0.0324} & \textbf{0.0168} & \textbf{0.0332} & \textbf{0.0677} \\
& SSIM$\uparrow$ & \textbf{0.9736} & \textbf{0.9675} & \textbf{0.9452} & \textbf{0.9661} & \textbf{0.9484} & \textbf{0.8953} \\
& DINO$\uparrow$ & \textbf{0.9974} & \textbf{0.9961} & \textbf{0.9924} & \textbf{0.9938} & \textbf{0.9889} & \textbf{0.9809} \\
& MSE$\downarrow$ & \textbf{0.0024} & \textbf{0.0033} & \textbf{0.0070} & \textbf{0.0047} & \textbf{0.0066} & \textbf{0.0146} \\
\bottomrule
\end{tabular}
\end{table*}

\section{Post-Processing: Color and Stroke Recovery}
\label{sec:postprocessing}

As described in the main paper, our model predicts colorless geometry from outline-based inputs. This design simplifies the learning task and improves robustness to appearance variations. Practical vectorization, however, also requires recovering colors and stroke properties from the original input image.
We present a comprehensive post-processing pipeline that leverages geometric congruence between predicted and input geometries to enable efficient color recovery, followed by z-order refinement and stroke detection.

\subsection{Geometric Congruence and Color Recovery}
\label{sec:color_extraction}

\paragraph{Training-Induced Geometric Alignment.}
A key property of our approach stems from the training data construction. 
During training, we ensure that the ground-truth SVG spatially aligns with the input outline raster image.
As a result, this geometric alignment carries over to inference: the model learns to predict rounded polygons that, when reconstructed to SVG, produce outlines spatially congruent with the input raster's underlying structure. 
This property enables efficient color recovery through overlap-based voting (as described below).

\paragraph{Overlap-Based Voting for Color Assignment.}
Given predicted paths $\mathcal{P} = \{P_1, \ldots, P_K\}$ and input raster $I_{\text{src}}$, we recover colors through a voting mechanism. A key challenge is that paths spatially overlap in the rendered output, making it ambiguous which input pixels should determine each path's fill color. To resolve this ambiguity, we must identify the pixels that exclusively correspond to each path—avoiding contamination from overlapping neighbors.

We determine this correspondence through a mask-based subtraction approach. For each path $P_i$, we construct a binary mask $M_i$ by rasterizing $P_i$ at its spatial position within the SVG coordinate frame (at resolution $r$, typically $256 \times 256$). Critically, all masks $\{M_1, \ldots, M_K\}$ are rasterized in the same coordinate space, preserving their relative positions—paths that are spatially distant will have non-intersecting masks, while overlapping paths will have intersecting masks. To isolate the exclusive pixel set for $P_i$, we compute the set difference: subtract the union of all other paths' masks from $M_i$. Formally, the exclusive mask is $M_i^{\text{excl}} = M_i \setminus \bigcup_{j \neq i} M_j$. The white pixels in $M_i^{\text{excl}}$ constitute the regions that uniquely belong to $P_i$ without being occluded by any other path. We then sample colors from $I_{\text{src}}$ at these exclusive pixel locations and assign the median (per RGB channel) as the fill color for $P_i$. 

However, when paths overlap heavily, some paths may have empty exclusive masks ($M_i^{\text{excl}} = \emptyset$). In practice, we address this using an iterative strategy: after each pass, paths that have been successfully assigned colors are removed from subsequent iterations. This progressively reveals previously occluded paths, allowing them to be colored in later passes. The process continues until all paths have been assigned colors or no further progress can be made. A possible extension would be to analyze the color distribution of pixels corresponding to each path in the raster input: if multiple peaks are present and the dominant color has already been assigned to a neighboring path in earlier iterations, selecting the next most significant unassigned peak may yield better results.

\subsection{Z-Order Optimization}
\label{sec:integrated_pipeline}

After the iterative color extraction process assigns colors to all paths using the initial ordering $\pi_0$ (the order in which polygons appear in the predicted representation), we perform z-order optimization to further improve reconstruction quality. The optimization searches for alternative orderings that minimize rendering error against $I_{\text{src}}$. The key steps are: (1) construct an overlap graph where paths with intersecting binary masks are connected by edges, (2) decompose this graph into independent connected components, (3) generate valid topological orderings within each component while respecting subset constraints, (4) prune the search space based on valid orderings, and (5) evaluate orderings via MSE comparison.

\subsection{Z-Order Optimization Details}
\label{sec:zorder}

\paragraph{Problem Formulation.}
Given initially colored paths $(\mathcal{P}, \mathcal{C}_0)$ in the initial ordering $\pi_0$, we seek the optimal ordering that minimizes rendering error:
\begin{equation}
\pi^* = \arg\min_{\pi} \text{MSE}(\text{Render}(\mathcal{P}, \mathcal{C}_0, \pi), I_{\text{src}})
\end{equation}
The optimization returns the single best ordering $\pi^*$ that achieves minimum MSE.

\paragraph{Connected Component Decomposition.}
We construct an overlap graph $G = (V, E)$ where each vertex represents a path and edges connect overlapping paths. Finding connected components via depth-first search decomposes the problem: paths in separate components can be optimized independently. This reduces complexity from $O(K!)$ to $O(\prod_j n_j!)$ where $n_j$ is the size of component $j$. For medium-complexity artworks, we found that most components have $n_j \leq 5$, making exhaustive enumeration tractable.

\paragraph{Topological Ordering with Pruning.}
Within each component, subset constraints form a directed acyclic graph (DAG): if $P_i \subset P_j$, then $P_j$ must render before $P_i$ (to avoid incorrect occlusion). We generate orderings via backtracking with topological constraints. Two key optimizations: (1) \emph{Independent pairs}: paths that don't overlap have commutative order—we select canonical representatives during generation to avoid redundant branches; (2) \emph{Early termination}: stop at $120$ evaluations and use area-based fallback (sort by area descending).

\paragraph{Evaluation and Selection.}
We optimize each component independently. For each component, we generate candidate orderings via the procedure described above (up to $120$ evaluations per component). To evaluate a component's orderings in isolation, we mask out all other components from $I_{\text{src}}$ and compute MSE only over the component's spatial region. This yields the optimal internal ordering for that component. Since paths in different components do not overlap, their relative inter-component order does not affect rendering. The final global ordering $\pi^*$ preserves the optimized intra-component orderings while allowing arbitrary arrangement of components relative to each other.

\subsection{Source-Guided Stroke Detection}
\label{sec:stroke_detection}

\paragraph{Motivation.}
Many vector graphics include strokes (outlines) in addition to fills. We employ an analytical method to determine stroke properties by directly measuring them from the input raster image.

\paragraph{Detection and Validation.}
For each path, we sample points along its geometry and compute the perpendicular (normal) direction at each point. We then traverse perpendicular to the path direction, stepping pixel-by-pixel until the color changes significantly. This directly measures the half-width of the stroke, and traversing in both normal directions ($+\mathbf{n}$ and $-\mathbf{n}$) yields the total stroke width. Individual measurements may vary due to anti-aliasing, path curvature, or sampling artifacts. One approach to handle this variability is to cluster width measurements and color measurements independently, then derive consensus values by selecting the median of the largest width cluster and the mode of the largest color cluster, which can provide robustness to outliers. To avoid false positives (e.g., detecting fills as strokes), we validate each detected stroke by rendering the SVG with the proposed stroke and computing pixel-level MSE against the source image. Strokes are accepted only if they improve MSE by at least $\epsilon = 0.0002$.

\paragraph{Pipeline Ordering.}
We perform fill color extraction and z-order optimization before stroke detection because, in most vector graphics, the majority of pixels correspond to path fills rather than strokes. By first optimizing fill colors and path ordering, which account for most of the reconstruction error, we substantially reduce MSE before adding strokes as a final refinement. In practice, this staged design is also more computationally efficient.

\paragraph{Practical Performance and Future Work.}
In our evaluation, this pipeline produces correct results for the vast majority of test cases while maintaining efficient computational cost (${\sim}200-300$ms per image on a single CPU core). The geometric congruence from training combined with iterative visibility-aware color extraction handles most occlusion scenarios effectively. However, these methods can fail on very complex SVG structures and cannot address all SVG properties (e.g., gradients, filters, transparency effects). Despite these limitations, we found them to work well in practice. Further improvement in handling complex structures and extending support for additional SVG properties remains an important direction for future work.

%% file: main.bib
@String(CVPR= {IEEE Conf. Comput. Vis. Pattern Recog.})

@String(ICCV= {Int. Conf. Comput. Vis.})

@String(TOG= {ACM Trans. Graph.})

@String(CVPR  = {CVPR})

@String(ICCV  = {ICCV})

@String(TOG   = {ACM TOG})

@article{Rodriguez_StarVector_2025,
  title={StarVector: Generating Scalable Vector Graphics Code from Images},
  author={Rodriguez, Juan A and others},
  journal={arXiv preprint arXiv:2312.11556},
  year={2025}
}

@article{yang2025omnisvg,
  title={OmniSVG: A Unified Scalable Vector Graphics Generation Model}, 
  author={Yiying Yang and Wei Cheng and Sijin Chen and Xianfang Zeng and Jiaxu Zhang and Liao Wang and Gang Yu and Xinjun Ma and Yu-Gang Jiang},
  journal={arXiv preprint arxiv:2504.06263},
  year={2025}
}

@article{xing2024llm4svg,
  title={Empowering LLMs to Understand and Generate Complex Vector Graphics},
  author={Xing, Ximing and Hu, Juncheng and Liang, Guotao and Zhang, Jing and Xu, Dong and Yu, Qian},
  journal={arXiv preprint arXiv:2412.11102},
  year={2024}
}

@article{Li_DiffVG_2020,
  title={Differentiable Vector Graphics Rasterization for Editing and Learning},
  author={Li, Tzu-Mao and Luk{\'a}{\v{c}}, Michal and Gharbi, Micha{\"e}l and Barron, Jonathan T},
  journal={ACM Transactions on Graphics (TOG)},
  volume={39},
  number={6},
  pages={1--15},
  year={2020},
  publisher={ACM}
}

@inproceedings{chen2024internvl,
  title={Internvl: Scaling up vision foundation models and aligning for generic visual-linguistic tasks},
  author={Chen, Zhe and Wu, Jiannan and Wang, Wenhai and Su, Weijie and Chen, Guo and Xing, Sen and Zhong, Muyan and Zhang, Qinglong and Zhu, Xizhou and Lu, Lewei and others},
  booktitle={Proceedings of the IEEE/CVF Conference on Computer Vision and Pattern Recognition},
  pages={24185--24198},
  year={2024}
}

@inproceedings{Caron_DINO_2021,
  title={Emerging Properties in Self-Supervised Vision Transformers},
  author={Caron, Mathilde and Touvron, Hugo and Misra, Ishan and J{\'e}gou, Herv{\'e} and Mairal, Julien and Bojanowski, Piotr and Joulin, Armand},
  booktitle={Proceedings of the IEEE/CVF International Conference on Computer Vision (ICCV)},
  pages={9650--9660},
  year={2021}
}

@inproceedings{Zhang_LPIPS_2018,
  title={The Unreasonable Effectiveness of Deep Features as a Perceptual Metric},
  author={Zhang, Richard and Isola, Phillip and Efros, Alexei A and Shechtman, Eli and Wang, Oliver},
  booktitle={Proceedings of the IEEE Conference on Computer Vision and Pattern Recognition (CVPR)},
  pages={586--595},
  year={2018}
}

@article{Wang_SSIM_2004,
  title={Image Quality Assessment: From Error Visibility to Structural Similarity},
  author={Wang, Zhou and Bovik, Alan C and Sheikh, Hamid R and Simoncelli, Eero P},
  journal={IEEE Transactions on Image Processing},
  volume={13},
  number={4},
  pages={600--612},
  year={2004},
  publisher={IEEE}
}

@article{Baran_Cornocopia_2010,
  title={Sketching Clothoid Splines Using Shortest Paths},
  author={Baran, Ilya and Lehtinen, Jaakko and Popovi\'{c}, Jovan},
  journal={Computer Graphics Forum},
  volume={29},
  number={2},
  pages={655--664},
  year={2010},
  doi={10.1111/j.1467-8659.2009.01635.x}
}

@misc{Potrace,
  title={Potrace: a polygon-based tracing algorithm},
  author={Selinger, Peter},
  year={2003},
  howpublished={\url{http://potrace.sourceforge.net/}}
}

@software{vtracer2020,
  author    = {Pun, Sanford and Tsang, Chris},
  title     = {{VTracer: Raster to Vector Graphics Converter}},
  year      = {2020},
  url       = {https://www.visioncortex.org/vtracer-docs},
  note      = {Vision Cortex documentation}
}

@article{ch25,
journal = {Computer Graphics Forum},
title = {{Image Vectorization via Gradient Reconstruction}},
author = {Chakraborty, Souymodip and Batra, Vineet and Phogat, Ankit and Jain, Vishwas and Ranawat, Jaswant Singh and Dhingra, Sumit and Wampler, Kevin and Fisher, Matthew and Lukác, Michal},
year = {2025},
publisher = {The Eurographics Association and John Wiley & Sons Ltd.},
ISSN = {1467-8659},
DOI = {10.1111/cgf.70055}
}

@article{wang2025internsvg,
  title={InternSVG: Towards Unified SVG Tasks with Multimodal Large Language Models},
  author={Wang, Haomin and Yin, Jinhui and Wei, Qi and Zeng, Wenguang and Gu, Lixin and Ye, Shenglong and Gao, Zhangwei and Wang, Yaohui and Zhang, Yanting and Li, Yuanqi and others},
  journal={arXiv preprint arXiv:2510.11341},
  year={2025}
}

@misc{chen2025svgeniusbenchmarkingllmssvg,
      title={SVGenius: Benchmarking LLMs in SVG Understanding, Editing and Generation}, 
      author={Siqi Chen and Xinyu Dong and Haolei Xu and Xingyu Wu and Fei Tang and Hang Zhang and Yuchen Yan and Linjuan Wu and Wenqi Zhang and Guiyang Hou and Yongliang Shen and          Weiming Lu and Yueting Zhuang},
      year={2025},
      eprint={2506.03139},
      archivePrefix={arXiv},
      primaryClass={cs.CV},
      url={https://arxiv.org/abs/2506.03139}, 
}

@inproceedings{xu2022live,
    title={Towards Layer-wise Image Vectorization},
    author={Ma, Xu and Zhou, Yuqian and Xu, Xingqian and Sun, Bin and Filev, Valerii and  Orlov, Nikita and Fu, Yun and Shi, Humphrey},
    booktitle={Proceedings of the IEEE conference on computer vision and pattern recognition},
    year={2022}
}

@software{cairosvg,
  author    = {Ayoub, Guillaume and {Kozea}},
  title     = {{CairoSVG: A Simple SVG Converter based on Cairo}},
  url       = {https://cairosvg.org/},
  note      = {Maintained by CourtBouillon},
  year      = {2012},
  publisher = {Kozea}
}

@software{skia,
  author    = {{Google LLC}},
  title     = {{Skia: A 2D Graphics Library}},
  url       = {https://skia.org/},
  note      = {An open source 2D graphics library sponsored and managed by Google.},
  year      = {2005},
  publisher = {Google LLC}
}

@techreport{openai2024gpt4o,
  title        = {GPT-4o System Card},
  author       = {OpenAI},
  institution  = {OpenAI},
  year         = {2024},
  month        = {August},
  url          = {https://cdn.openai.com/gpt-4o-system-card.pdf}
}

@article{gemini2025gemini2_5,
  title        = {Gemini 2.5: Pushing the Frontier with Advanced Reasoning, Multimodality, Long Context, and Next Generation Agentic Capabilities},
  author       = {{Gemini Team, Google}},
  journal      = {arXiv preprint arXiv:2507.06261},
  year         = {2025},
  eprint       = {2507.06261},
  archiveprefix = {arXiv},
  url          = {https://arxiv.org/abs/2507.06261}
}

@inproceedings{Yang2016EffectiveClipart,
  author    = {Ming Yang and Hongyang Chao and Chi Zhang and Jun Guo and Lu Yuan and Jian Sun},
  title     = {Effective Clipart Image Vectorization Through Direct Optimization of Bezigons},
  booktitle = {arXiv preprint arXiv:1602.01913},
  year      = {2016},
  note      = {Available online},
  url       = {https://arxiv.org/abs/1602.01913}
}

@misc{hu2021lora,
  title        = {LoRA: Low-Rank Adaptation of Large Language Models},
  author       = {Edward J. Hu and Yelong Shen and Phillip Wallis and Zeyuan Allen-Zhu and Yuanzhi Li and Shean Wang and Lu Wang and Weizhu Chen},
  year         = {2021},
  howpublished = {arXiv preprint arXiv:2106.09685},
  eprint       = {2106.09685},
  archivePrefix= {arXiv},
  primaryClass = {cs.CL},
  url          = {https://arxiv.org/abs/2106.09685}
}

@inproceedings{loshchilov2019adamw,
  title     = {Decoupled Weight Decay Regularization},
  author    = {Loshchilov, Ilya and Hutter, Frank},
  booktitle = {International Conference on Learning Representations},
  year      = {2019},
  url       = {https://arxiv.org/abs/1711.05101}
}

@misc{snell2024scaling,
  title        = {Scaling LLM Test-Time Compute Optimally Can Be More Effective than Scaling Model Parameters},
  author       = {Charlie Snell and Jaehoon Lee and Kelvin Xu and Aviral Kumar},
  year         = {2024},
  howpublished = {arXiv preprint arXiv:2408.03314},
  eprint       = {2408.03314},
  primaryClass = {cs.CL},
  url          = {https://arxiv.org/abs/2408.03314}
}

@misc{rodriguez2025renderingawarereinforcementlearningvector,
      title={Rendering-Aware Reinforcement Learning for Vector Graphics Generation}, 
      author={Juan A. Rodriguez and Haotian Zhang and Abhay Puri and Aarash Feizi and Rishav Pramanik and Pascal Wichmann and Arnab Mondal and Mohammad Reza Samsami and Rabiul Awal and Perouz Taslakian and Spandana Gella and Sai Rajeswar and David Vazquez and Christopher Pal and Marco Pedersoli},
      year={2025},
      eprint={2505.20793},
      archivePrefix={arXiv},
      primaryClass={cs.CV},
      url={https://arxiv.org/abs/2505.20793}, 
}

@manual{adobeillustrator2025,
  title        = {Adobe Illustrator},
  author       = {{Adobe Inc.}},
  organization = {Adobe Systems Incorporated},
  address      = {San Jose, California, USA},
  year         = {2025},
  note         = {Version 29.1 (Creative Cloud)},
  url          = {https://www.adobe.com/products/illustrator.html}
}
